\documentclass[11pt]{article}
\usepackage[a4paper, margin=1in]{geometry}
\usepackage{iftex}
\ifPDFTeX
  \usepackage[T1]{fontenc}
  \usepackage[utf8]{inputenc}
  \usepackage{lmodern}
  \usepackage{textcomp}
\else
  \usepackage{fontspec}
\fi
\usepackage{amsmath,amssymb}
\usepackage[round]{natbib}
\usepackage{graphicx}
\usepackage{booktabs}
\usepackage{tabularx}
\usepackage{rotating}
\usepackage{adjustbox}
\usepackage{tikz}
\usetikzlibrary{arrows.meta,calc,positioning}
\usepackage{enumitem}
\usepackage{microtype}
\usepackage{xcolor}
\usepackage[colorlinks=true, linkcolor=blue!60!black, citecolor=blue!60!black, urlcolor=blue!60!black]{hyperref}
\usepackage{caption}
\hypersetup{pdftitle={Reification as a Transferable Vocabulary: Zero-Shot Link Prediction with Vanilla GNNs},
  pdfauthor={Camille Pradel}}
\usepackage[section]{placeins}
\usepackage{chngcntr}
\providecommand{\tightlist}{\setlength{\itemsep}{0pt}\setlength{\parskip}{0pt}}
\setlist{itemsep=2pt, topsep=4pt}

\title{Reification as a Transferable Vocabulary: Zero-Shot Link Prediction with Vanilla GNNs}

\author{Camille Pradel\\
  Matr, \url{https://itmatr.tech}\\
  \texttt{camille@itmatr.tech}}
\date{}

\begin{document}
\maketitle

\begin{abstract}
Knowledge graph foundation models such as ULTRA achieve zero-shot link
prediction on unseen graphs through dedicated architectures that hard-code
a transfer mechanism. In this work we move that mechanism out of the
architecture and into the representation, by \emph{reifying} the input graph:
every fact becomes a node, connected to its subject, object, and
relation type through a fixed vocabulary of six meta-relations, with
relation types as anonymous shared nodes rather than model parameters.
On this representation, five textbook GNNs (GAT, GINE with sum and with
mean+max aggregation, GraphSAGE, R-GCN), each trained on a
single knowledge graph of 4,245 triples for 30 minutes on one NVIDIA
A100, transfer zero-shot to 40 inductive link-prediction benchmarks.
The best of them, an off-the-shelf GAT, matches ULTRA, a dedicated
foundation model pretrained on three graphs, across ULTRA's own
evaluation suite. The same fixed
vocabulary extends to relational databases, a row becoming an entity
and a foreign-key column a relation type; a preliminary probe on two
unseen databases, with no cell values, schema text or in-context labels,
shows a model of this family pretrained on three knowledge graphs
ranking foreign-key targets far above random-initialization and degree
controls. We release the
code, the checkpoints, and the evaluation pipeline for all 40
benchmarks.
\end{abstract}

\section{Introduction}

The foundation-model recipe (pre-train on broad data, apply zero-shot to
unseen inputs) rests on a shared input vocabulary. A language model can
read any document because every document is written in the same tokens.
Relational data offers no such shared vocabulary. Every knowledge graph
names its own entities and its own relation types; every relational
database adds one more private layer: a schema of tables and columns
found nowhere else, its foreign keys playing the role of a knowledge
graph's relation types. Any model that allocates
parameters to a vocabulary item (an entity embedding, a relation embedding,
a per-table encoder) is structurally unable to read the next dataset.
Recent position work on graph foundation models identifies exactly this as
the central open problem: finding a \emph{transferable vocabulary} shared across
graphs \citep{mao2024position}.

Two research lines answer it today, each confined to one data model and
each with a dedicated architecture. On the knowledge graph side,
\emph{knowledge graph foundation models} (KGFMs) achieve structural transfer:
ULTRA \citep{galkin2024foundationmodelsknowledgegraph} conditions on the query
and learns relative relation representations built from a relation graph
with four fixed relation-to-relation interactions; TRIX, MOTIF, and Gamma
refine the same design \citep{zhang2025trixexpressivemodelzeroshot, huang2025expressiveknowledgegraphfoundation, xin2025geometricstructuralknowledgegraph}; Flock replaces message
passing with random-walk sequence encoding \citep{kim2025flock}; prompt- and
in-context variants condition a pretrained reasoner with graph prompts
\citep{cui2024kgicl, wang2024prolink}, and SEMMA adds text embeddings of
relation names \citep{arun2025semmasemanticawareknowledge}. These models
generalize to unseen entities and relations, but they are KG-only, and
each hard-codes its transfer mechanism into a bespoke architecture. On
the relational database side, \emph{relational foundation models} transfer
through a mix of schema text and in-context conditioning: the Relational
Transformer \citep{ranjan2026relationaltransformerzeroshotfoundation}
tokenizes every cell with its column name and value text, while KumoRFM
\citep{hudovernik2026kumorfm2scalingfoundationmodels} learns in-context over a
relational graph transformer. These generalize to unseen schemas, but
they are RDB-only. A knowledge graph can certainly be stored as a
database, as one triple table or as one table per relation type, and its
relation names then provide a thin layer of schema text; what the
standard link-prediction benchmarks cannot provide is the labeled task
table that carries most of these models' transfer (Section 2). No single
model covers both, and in every one of them the
transfer mechanism lives in the model.

Our idea is to move the problem from the architecture to the
representation. We \emph{reify} the data (Figure 1): every fact becomes a
node, connected to its subject, its object, and its relation type through
a fixed vocabulary of six meta-relations. Relation types become
\emph{anonymous shared nodes} rather than model parameters, and so do entity
types (tables) in the database case. The nodes are shared in that two
facts of the same relation point to the same relation-type node, so
``being of the same kind'' is now structure in the graph rather than a row
in an embedding table. They are anonymous in that they carry no learned
identity: the model reads nothing but a node's kind and its structural
position. After reification, every dataset is a homogeneous graph over
the same finite input vocabulary: three node kinds (entities, facts,
relation types), four with entity types for databases, and six
meta-relations, eight with \texttt{is\_a} and its inverse (Section 3.1). Nothing
dataset-specific remains for a model to parameterize. Any GNN that is
inductive over nodes, given query conditioning implemented as a labeling
trick \citep{zhang2022labelingtricktheoryusing}, can be trained on one dataset
and applied unchanged to any other.

The question is then empirical: with the representation fixed and the
architecture deliberately generic, how much zero-shot inductive transfer
is left? We report two findings and one probe.

\begin{enumerate}
\def\labelenumi{\arabic{enumi}.}
\tightlist
\item
  \textbf{Zero-shot transfer works with unmodified architectures.} Five
  textbook backbones (GAT \citep{velickovic2018graphattention}, GINE with sum
  and with mean+max aggregation \citep{xu2019powerful, hu2020strategies}, GraphSAGE
  \citep{hamilton2017inductive}, and R-GCN \citep{schlichtkrull2018modeling}),
  trained under one frozen recipe (64 dimensions, 12 layers, a
  30-minute budget on one NVIDIA A100, 3 seeds fixed a priori) on a
  single 4,245-triple knowledge
  graph, all transfer zero-shot to 40 inductive benchmarks. None
  collapses, with one exception, identified and explained
  (Section 5.3), and the ordering between them is stable across
  evaluation panels up to one swap (Section 5.1).
\item
  \textbf{The best of them matches the dedicated foundation model.} Over all
  40 benchmarks the GAT averages 0.3564 ± 0.0037 against 0.3732 for
  ULTRA re-run from its released checkpoint with our evaluation code;
  on the 13 benchmarks whose families appear in neither pretraining
  corpus, 0.3609 ± 0.0062 against 0.3617; and on the twelve
  inductive-(e) GraIL benchmarks it scores higher, 0.5457 against
  0.5224 (Section 5.2).
\item
  \textbf{A probe of the same representation on relational databases.} A
  model of this family pretrained on three knowledge graphs, applied to
  two unseen reified databases with no features, no fine-tuning, no
  schema text and no in-context labels, ranks foreign-key targets far
  above random-initialization and degree controls, while the
  single-graph models of Section 5 do not (Section 6). This is where the
  vocabulary's coverage of both data models is put to the test; we
  report the outcome as a preliminary probe, not as a transfer result.
\end{enumerate}

We do not claim to beat the state of the art: ULTRA remains ahead on the
full aggregate, the KGFMs that refine ULTRA's design (TRIX, MOTIF, Gamma,
Flock) publish figures up to 0.03 MRR above ULTRA's on the inductive-(e)
splits (Table 5), and Section 5.4 shows the remaining gap is not closed
by choosing a different single training graph. The contribution is the
demonstration that a fixed structural representation carries most of
zero-shot inductive transfer with none of the dedicated machinery. The
representation itself is not new as an encoding: the literature uses
reification for n-ary facts, and HYPER
\citep{huang2026hyperfoundationmodelinductive} treats it as a hostile
baseline, showing that ULTRA pretrained on ordinary KGs transfers poorly
to reified graphs. To our knowledge, no prior work pre-trains natively on
reification or proposes it as the cross-dataset transfer mechanism
(Section 2).

\begin{figure}[t]
\centering
\resizebox{\textwidth}{!}{\begin{tikzpicture}[
  font=\scriptsize,
  ent/.style={draw=black!60, rounded corners=2pt, fill=black!4, inner sep=3pt, minimum height=15pt, align=center},
  fact/.style={draw=black!70, circle, fill=blue!8, inner sep=1.2pt, align=center},
  rel/.style={draw=orange!80!black, rounded corners=2pt, fill=orange!20, inner sep=3pt, minimum height=15pt, align=center},
  cls/.style={draw=green!50!black, rounded corners=2pt, fill=green!12, inner sep=3pt, minimum height=15pt, align=center},
  e/.style={-{Stealth[length=4.5pt]}, black!70, thin},
  lab/.style={font=\tiny, fill=white, inner sep=1pt, sloped, midway},
  panelname/.style={font=\scriptsize\bfseries, anchor=north},
  tbl/.style={font=\tiny, inner sep=0pt},
  every node/.style={align=center}
]
\begin{scope}[local bounding box=panelA]
  \node[ent] (paris)  at (0,0)     {Paris};
  \node[ent] (france) at (2.4,0)   {France};
  \node[ent] (madrid) at (0,-1.5)  {Madrid};
  \node[ent] (spain)  at (2.4,-1.5){Spain};
  \draw[e] (paris)  -- node[lab, above]{capital\_of} (france);
  \draw[e] (madrid) -- node[lab, above]{capital\_of} (spain);
\end{scope}
\node[panelname] at (1.2,-2.3) {(a) two facts of a knowledge graph};

\begin{scope}[shift={(4.6,0)}, local bounding box=panelB]
  \node[rel] (r) at (2.7,1.7) {REL\_TYPE \#42\\[-2pt]\tiny\itshape (capital\_of)};
  \node[ent]  (p)  at (0,0)     {ENTITY \#3\\[-2pt]\tiny\itshape (Paris)};
  \node[fact] (f1) at (2.7,0)   {\tiny FACT\\[-3pt]\tiny \#17};
  \node[ent]  (fr) at (5.4,0)   {ENTITY \#6\\[-2pt]\tiny\itshape (France)};
  \node[ent]  (m)  at (0,-1.5)  {ENTITY \#4\\[-2pt]\tiny\itshape (Madrid)};
  \node[fact] (f2) at (2.7,-1.5){\tiny FACT\\[-3pt]\tiny \#18};
  \node[ent]  (sp) at (5.4,-1.5){ENTITY \#9\\[-2pt]\tiny\itshape (Spain)};
  \draw[e] (f1) -- node[lab, above]{has\_subject} (p);
  \draw[e] (f1) -- node[lab, above, pos=0.55]{has\_object} (fr);
  \draw[e] (f2) -- node[lab, above]{has\_subject} (m);
  \draw[e] (f2) -- node[lab, above, pos=0.68]{has\_object} (sp);
  \draw[e, orange!80!black] (f1) -- node[lab, right, sloped=false, pos=0.45]{has\_type} (r);
  \draw[e, orange!80!black] (f2) to[bend right=22] node[lab, sloped=false, right, pos=0.28]{has\_type} (r.east);
\end{scope}
\node[panelname] at (7.3,-2.3) {(b) the same facts, reified};

\begin{scope}[shift={(0,-5.5)}, local bounding box=panelC]
  \node[tbl, anchor=north west] (t1) at (-0.7,1.9) {%
    \textbf{results}\\[1pt]
    \begin{tabular}{@{}c c c c@{}}
      \toprule resultId & raceId & driverId & $\cdots$ \\ \midrule
      88 & 12 & \textbf{14} & $\cdots$ \\
      89 & 13 & \textbf{14} & $\cdots$ \\
      $\vdots$ & $\vdots$ & $\vdots$ & \\ \bottomrule
    \end{tabular}};
  \node[tbl, anchor=north west] (t2) at ([yshift=-0.3cm]t1.south west) {%
    \textbf{drivers}\\[1pt]
    \begin{tabular}{@{}c c c@{}}
      \toprule driverId & name & $\cdots$ \\ \midrule
      \textbf{14} & Alonso & $\cdots$ \\
      $\vdots$ & $\vdots$ & \\ \bottomrule
    \end{tabular}};
\end{scope}
\node[panelname] at (1.2,-8.15) {(c) a database extract};

\begin{scope}[shift={(5.0,-5.5)}, local bounding box=panelD]
  \node[cls] (c1) at (0,1.7)   {CLASS \#1\\[-2pt]\tiny\itshape (results)};
  \node[rel] (fk) at (2.7,1.7) {REL\_TYPE \#7\\[-2pt]\tiny\itshape (results.driverId)};
  \node[cls] (c2) at (5.4,1.7) {CLASS \#2\\[-2pt]\tiny\itshape (drivers)};
  \node[ent]  (r88) at (0,0)     {ENTITY \#88\\[-2pt]\tiny\itshape (results row 88)};
  \node[fact] (g1)  at (2.7,0)   {\tiny FACT\\[-3pt]\tiny \#91};
  \node[ent]  (d14) at (5.4,0)   {ENTITY \#14\\[-2pt]\tiny\itshape (drivers row 14)};
  \node[ent]  (r89) at (0,-1.5)  {ENTITY \#89\\[-2pt]\tiny\itshape (results row 89)};
  \node[fact] (g2)  at (2.7,-1.5){\tiny FACT\\[-3pt]\tiny \#92};
  \draw[e] (g1) -- node[lab, above]{has\_subject} (r88);
  \draw[e] (g1) -- node[lab, above, pos=0.55]{has\_object} (d14);
  \draw[e] (g2) -- node[lab, above]{has\_subject} (r89);
  \draw[e] (g2) -- node[lab, below, pos=0.5]{has\_object} (d14);
  \draw[e, orange!80!black] (g1) -- node[lab, right, sloped=false, pos=0.45]{has\_type} (fk);
  \draw[e, orange!80!black] (g2) to[bend right=22] node[lab, sloped=false, right, pos=0.28]{has\_type} (fk.east);
  \draw[e, green!50!black] (r88) -- node[lab, left, sloped=false, pos=0.5]{is\_a} (c1);
  \draw[e, green!50!black] (r89) to[bend left=35] node[lab, sloped=false, left, pos=0.5]{is\_a} (c1.west);
  \draw[e, green!50!black] (d14) -- node[lab, right, sloped=false, pos=0.5]{is\_a} (c2);
\end{scope}
\node[panelname] at (7.7,-8.15) {(d) the extract, reified};
\end{tikzpicture}}
\caption{\textbf{Reification as a fixed structural vocabulary.} Panels (a) and (b): two facts of a knowledge graph and their reification. Panels (c) and (d): an extract of a relational database, two tables linked by one foreign key, and its reification into the same vocabulary. Inverse meta-relations are omitted for readability; annotations in parentheses are for the reader, the model sees only anonymous node identities and kinds.}

\end{figure}
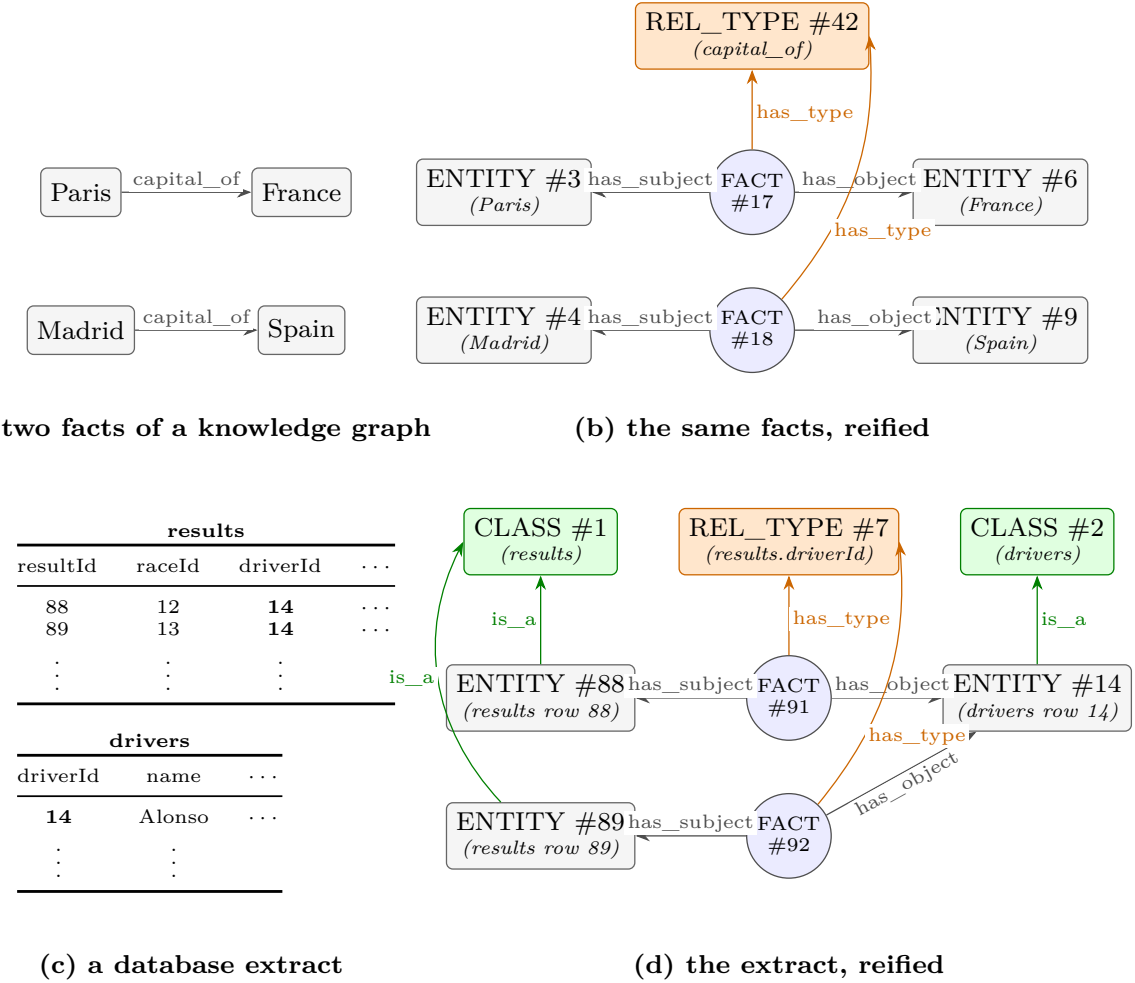

\section{Related work}

\textbf{Inductive link prediction and KG foundation models.} GraIL
\citep{teru2020inductiverelationpredictionsubgraph} introduced inductive
relation prediction over unseen entities (the inductive-(e) scenario) via
subgraph reasoning; NBFNet \citep{zhu2022neuralbellmanfordnetworksgeneral}
made source-conditioned full-graph propagation the dominant backbone;
A*Net and AdaProp address its cost \citep{zhu2023anet, zhang2023adaprop}.
InGram \citep{lee2023ingraminductiveknowledgegraph} first targeted unseen
\emph{relations} (inductive-(e,r)) through a relation graph. ULTRA
\citep{galkin2024foundationmodelsknowledgegraph} turned that relation-graph
idea into the first KGFM: relative relation representations from a
relation graph with four fundamental interactions (h2h, h2t, t2h, t2t),
conditioned on the query, in two stacked NBFNet-style conditional MPNNs.
TRIX \citep{zhang2025trixexpressivemodelzeroshot} adds recursive
entity/relation updates with provably higher expressiveness; MOTIF
\citep{huang2025expressiveknowledgegraphfoundation} characterizes what KGFMs
based on binary relation motifs can express; Gamma
\citep{xin2025geometricstructuralknowledgegraph} adds multi-geometry
attention; Flock \citep{kim2025flock} breaks with this design, replacing
message passing with random-walk sequence encoding. On the GraIL and
InGram splits, TRIX and Flock publish the highest zero-shot figures of
the four (Section 5.2, Table 5). Prompt- and in-context KGFMs
\citep{cui2024kgicl, wang2024prolink, gao2026kgpfnunlockingpotentialknowledge}
condition a pretrained reasoner with graph prompts; SEMMA
\citep{arun2025semmasemanticawareknowledge} adds LLM text embeddings of
relation names. Apart from Flock and the in-context line, these models
are direct descendants of ULTRA's relation-graph design; ULTRA remains
the family's shared reference point, and our evaluation uses it as such
(Section 4). All of these are KG-only, dedicated architectures: the
transfer mechanism lives in the model. We put it in the data. ULTRA's
relation graph is a structural transformation too, but only ULTRA can
consume it; the reified graph can be read by any node-inductive GNN, and
this paper is that substitution test.

\textbf{Relational deep learning and RDB foundation models.} Relational Deep
Learning \citep{fey2023relationaldeeplearninggraph} reads a relational
database as a heterogeneous temporal graph, the \emph{relational entity
graph}; RelBench \citep{robinson2024relbenchbenchmarkdeeplearning} is its
benchmark suite. Two foundation models transfer
across databases: the Relational Transformer (RT)
\citep{ranjan2026relationaltransformerzeroshotfoundation}, which tokenizes
every cell with its column name and value and pre-trains on masked
cells, and KumoRFM \citep{hudovernik2026kumorfm2scalingfoundationmodels}, its
commercial counterpart, which learns in-context over a relational graph
transformer. What carries their zero-shot transfer is best read from
RT's own context ablation: in-context labels of the target entity first,
schema text second; and since column-name overlap between RelBench
databases is near zero, RT concludes it must also exploit structural
relational patterns
\citep{ranjan2026relationaltransformerzeroshotfoundation}.

None of these mechanisms is available on a bare knowledge graph. A KG
can be stored as a database, as one triple table or as one table per
relation type, and its relation names then provide a thin layer of
schema text; but the link-prediction benchmarks provide no labeled task
table to condition on, and that is the dominant channel. The gap is also
one of task: RT's pre-training predicts value cells, foreign keys shape
its attention but are never prediction targets, and its authors list
link prediction as out of scope, so RT can appear neither in the KG
evaluation of Section 5 nor in the featureless probe of Section 6. We
therefore describe each family by its mix of three composable transfer
mechanisms: structural invariants, schema text, and in-context
conditioning on labeled examples (Table 1). Our representation uses the
first alone: at inference it consumes the target dataset and nothing
else.

\begin{table}[t]
\centering
\footnotesize
\caption{\textbf{positioning by transfer mechanism} (S = structural invariants, T = schema/relation-name text, C = in-context conditioning on labeled examples).}

\begin{tabularx}{\textwidth}{>{\raggedright\arraybackslash}p{3.3cm}>{\raggedright\arraybackslash}X>{\raggedright\arraybackslash}X>{\raggedright\arraybackslash}X>{\raggedright\arraybackslash}X}
\toprule
 & mechanism(s) & domains & architecture & generalizes to \\
\midrule
ULTRA / TRIX / MOTIF / Gamma / Flock & S (relation graph / walks) & KGs & dedicated & unseen entities + relations \\
SEMMA / ProLINK & S + T (relation-name text via LLM) & KGs & dedicated & unseen entities + relations \\
KG-ICL / KGPFN & S + C (prompt graphs / episodes) & KGs & dedicated & unseen entities + relations \\
RT / KumoRFM & C (dominant, per RT's ablation) + T + learned S & RDBs & dedicated & unseen entities + relations + schemas \\
\textbf{This work} & \textbf{S, in the representation} & \textbf{KGs (+ RDB probe)} & \textbf{any inductive MPNN} & \textbf{unseen entities + relations + schemas} \\
\bottomrule
\end{tabularx}
\end{table}

\textbf{Reification.} Turning an edge into a node is classical: the Levi graph
in graph theory and star expansion for hypergraphs \citep{agarwal2006higher}.
Message passing centred on edges rather than on entities has been used
as an architectural device for link prediction: PathCon
\citep{wang2021pathcon} aggregates the relational context and the relational
paths around an entity pair without any entity embedding, and RMPI
\citep{geng2023rmpi} propagates between the edges of an enclosing subgraph,
on its line graph, to reach unseen relations as well as unseen entities,
drawing relation semantics from an ontological schema when one is
available. Both keep the edge-centric view inside a dedicated model with
relation-specific parameters; here the same view is written into the
data once, with relation types as anonymous nodes, so that a homogeneous
GNN with no relation parameters can read it. In the KG literature, reification is the standard encoding for n-ary and
hyper-relational facts \citep{wei2025narysurvey}, and the hyper-relational
line explicitly engineered \emph{around} it for performance: StarE
\citep{galkin2020stare} and HINGE \citep{rosso2020hinge} keep qualifiers native;
HYPER \citep{huang2026hyperfoundationmodelinductive} operates directly on
hypergraphs and evaluates ``ULTRA over reification'' as a degraded
baseline, attributing the drop to longer hop distances and distribution
shift from ordinary-KG pre-training. We draw the opposite conclusion from
the same observation: reification hurts models pre-trained on
\emph{non-reified} data, and, to our knowledge, nobody has pre-trained
natively on the reified representation. Concurrently,
\citep{amouzouvi2026graphlets} also seek a universal structural vocabulary for
KGFMs, via graphlet \emph{features} fed to a dedicated model; ours is a
representational substrate: the data itself is rewritten.

\textbf{Expressiveness and equivariance.} Unconditioned GNNs compute unary node invariants and provably
cannot solve link prediction; conditioning on the query (labeling trick
\citep{zhang2022labelingtricktheoryusing}, C-MPNN
\citep{huang2023theorylinkpredictionrelational}) is necessary. Double
equivariance \citep{zhou2025doubleequivarianceinductivelink} formalizes the
joint entity$\times$relation invariance that inductive-(e,r) prediction
requires, and reification realizes it structurally: permuting relation
types permutes anonymous nodes, so any node-permutation-equivariant GNN
is automatically equivariant to it. ISDEA, the model of that work,
obtains the same invariance through a dedicated architecture; its
evaluation ranks each answer against 50 sampled negatives, so its
published figures are not comparable with the filtered full-ranking
protocol used throughout this paper, and we do not compare against them
numerically. Chen et al.
\citep{chen2023calibrateboostlogicalexpressiveness} analyze how graph
transformations trade expressiveness against depth, the source of our
``one original hop = two reified hops'' accounting.

\textbf{Broader structured-data foundation models.} TabPFN
\citep{hollmann2023tabpfntransformersolvessmall} and TabICL
\citep{qu2025tabicltabularfoundationmodel} cover single tables; GraphAny
\citep{zhao2024graphany} targets node classification on arbitrary graphs;
Wang et al. \citep{wang2025graphfoundationmodelstraining} transfer from KGs to
general graphs via textualization, the closest prior to our database
probe, with text rather than structure as the bridge.

\section{Method}

\subsection{The reified representation}

\textbf{Knowledge graphs.} A KG is a triple store \(G=(E,R,F)\) with entities
\(E\), relation types \(R\), and facts \(F \subseteq E \times R \times E\). Its
input vocabulary, what a model must be able to read, is \(E \cup R\), and
both parts change with every dataset.

\textbf{Reification.} We map \(G\) to a homogeneous directed graph \(G'=(V',E')\)
(Figure 1a--b) with:

\begin{itemize}
\tightlist
\item
  \textbf{Nodes} \(V' = E \,\cup\, F \,\cup\, R\): one node per entity
  (kind \texttt{ENTITY}), per fact (kind \texttt{FACT}), and per relation type
  (kind \texttt{REL\_TYPE}). Relation-type nodes are \emph{anonymous}: they carry no
  identity beyond their kind and their structural position.
\item
  \textbf{Edges}: for each fact \(f=(h,r,t)\), six edges over a fixed
  meta-relation vocabulary: \texttt{has\_subject} \((f{\to}h)\), \texttt{subject\_of}
  \((h{\to}f)\), \texttt{has\_object} \((f{\to}t)\), \texttt{object\_of} \((t{\to}f)\),
  \texttt{has\_type} \((f{\to}r)\), \texttt{has\_instance} \((r{\to}f)\).
\end{itemize}

The direction of the original relation is preserved \emph{structurally}
(\texttt{has\_subject} vs \texttt{has\_object}), so inverse-relation augmentation, a
staple of multi-relational GNNs, is unnecessary and omitted. The input
vocabulary of \(G'\) is finite and universal: 3 node kinds and 6
meta-relations, whatever the dataset. Two facts of the same relation type
share a relation node; ``these two edges are of the same kind'' is a 2-hop
structural pattern, not a parameter.

\textbf{Relational databases.} A relational database maps into the same
vocabulary, extended by exactly one node kind and one meta-relation. The
mapping has two steps. \emph{Step 1: the database is already a
multi-relational graph.} Following relational deep learning
\citep{fey2023relationaldeeplearninggraph}, take every row of every table as a
node, and every foreign-key cell (a cell of row \(u\) holding the key of a
row \(v\) in another table) as an edge \(u \to v\). Two edges coming from
the same foreign-key column are of the same type, exactly as two KG facts
sharing a relation: the foreign-key columns \emph{are} the database's relation
types. \emph{Step 2: reify that graph exactly as above.} Each foreign-key cell
becomes a \texttt{FACT} node with its six meta-relation edges, and each
foreign-key column an anonymous \texttt{REL\_TYPE} node. The one genuinely new
ingredient is the schema's type information: each table becomes an
entity-type node of a fourth kind, \texttt{CLASS}, connected to its rows by one
more meta-relation pair, \texttt{is\_a} and its inverse \texttt{has\_member} (Figure 1c--d).
A new database is therefore new structure over the same fixed vocabulary,
exactly like a new KG. The models of this paper are pretrained on
knowledge graphs, which have no tables, so they never see a class node;
Section 6 says how the probe accommodates this.

\textbf{Remark (\texttt{has\_type} vs \texttt{is\_a}).} Both are instantiation edges
(instance $\rightarrow$ type) and could in principle be merged into one
meta-relation, the endpoint node kinds carrying the distinction. We keep
them separate, mirroring RDF's \texttt{rdf:predicate} / \texttt{rdf:type} distinction:
\texttt{has\_type} sits on the critical path of query conditioning and relational
composition (every fact has exactly one), while \texttt{is\_a} is an optional
schema signal that a bare KG does not have. Distinct edge embeddings let
the model weight the two roles independently from the first layer.

\textbf{Structure versus data properties.} The vocabulary covers structure
only: entities (rows), relation types (foreign-key columns), facts
(foreign-key cells), and entity types (tables). Data properties, the
non-key cell values, are not part of the reified graph; integrating
them is left to future work. The restriction is what allows the database probe
of Section 6 to test the vocabulary itself: run featureless, it
measures what key structure alone carries, not the quality of a feature
encoder. The KG benchmarks of Section 5 contain no literals, so the
restriction costs them nothing.

\textbf{What reification costs.} It multiplies node count (roughly $\times$20 on
typical KGs) and doubles the graph diameter, since one original hop
becomes two reified hops through the fact node. Section 3.2 sets depth
accordingly and Section 7 discusses the cost.

\textbf{What reification buys.} It reduces the requirement on the model from
``multi-relational GNN with a parametric relation vocabulary'' to
``homogeneous GNN inductive over nodes''. It realizes the entity$\times$relation double
anonymization that inductive-(e,r) prediction demands
\citep{zhou2025doubleequivarianceinductivelink} structurally: permuting
relation types permutes anonymous nodes, and any
node-permutation-equivariant GNN is automatically equivariant to it.
The correspondence extends to computation: one layer of a conditional
multi-relational MPNN on \(G\) aggregates over typed edges, and the same
aggregation factors through the fact nodes of \(G'\) in two hops, so \(k\)
layers of a query-conditioned homogeneous MPNN on \(G'\) can carry out
\(\lfloor k/2 \rfloor\) layers of message passing on \(G\). We use this as
a depth accounting (Section 3.2) rather than as a formal result.

\subsection{The five backbones}

\textbf{Why conditioning is necessary.} An unconditioned MPNN computes unary
node invariants, which provably cannot rank links
\citep{zhang2022labelingtricktheoryusing, huang2023theorylinkpredictionrelational}. We therefore condition the
\emph{initialization} on the query: the minimal, architecture-agnostic form of
the labeling trick.

\textbf{Initialization.} Node states start from their kind embedding,
\(x_v = \mathrm{emb}_{\mathrm{kind}}(\mathrm{kind}(v))\) (3 vectors). For a
query \((h, r, ?)\) (resp. \((?, r, t)\)), two learned additive markers are
applied: a \emph{role} marker on the known endpoint entity (subject vs object
role; 2 vectors) and a \emph{query-relation} marker on the queried
relation-type node (1 vector). Propagation then runs on one copy of the
graph per query. Total conditioning parameters: 3 vectors of dimension
\(d\). There are no per-relation, per-entity, or per-dataset parameters
anywhere in the model.

\textbf{Backbones.} The representation can be consumed by any MPNN that is
inductive over nodes; only per-entity shallow embeddings, transductive
by construction, are out of scope. To keep the architecture out of the
result, we take four backbones exactly as shipped by PyTorch Geometric
\citep{fey2019fast} (GAT, GINE with sum aggregation, GraphSAGE, and R-GCN over
the six meta-relations) and one two-line variant of GINE whose sum
aggregator is replaced by PyG's mean+max multi-aggregation; we change
nothing else in them. GraphSAGE, as shipped, consumes no edge features
and therefore does not see the meta-relation labels; it is kept as the
weakest-information baseline. All five share one configuration: dimension 64, 12 layers ($\approx$6 logical hops
on the original graph, given the $\times$2 diameter), LayerNorm and dropout 0.2
per layer, and 0.11M--0.41M parameters depending on backbone, the same
order as ULTRA's 0.17M.

\textbf{Scoring, and the asymmetry it creates.} Candidate \(t\) is scored by an
MLP over its final state alone, \(s(t) = \mathrm{MLP}(x_t)\). This is the
weakest readout available, chosen so that the representation carries the
result. It leaves the models query-conditioned at the \emph{propagation} level
(through seeded initialization) but not at the \emph{readout} level, where
ULTRA also conditions its scoring on the query. Section 5.3 shows that
this asymmetry is invisible on 39 of the 40 benchmarks and decisive on
the fortieth.

\textbf{Training.} Cross-entropy over all entities (full softmax), with known
true answers masked (filtered training). To prevent leakage, the queried fact's six
edges (its three meta-relation edges and their inverses, Section 3.1)
are removed from its graph copy during message passing.
Training and inference are full-graph, NBFNet-style
\citep{zhu2022neuralbellmanfordnetworksgeneral}. The recipe was frozen before
any evaluation: 30-minute budget on one NVIDIA A100, selection on
validation MRR only, seeds \{0, 1, 2\} fixed a priori for every backbone,
no post-hoc seed or checkpoint selection.

\section{Experimental setup}

\textbf{Training data.} Every model is trained on a single graph: the
training split of FB15k-237-Inductive v1
\citep{teru2020inductiverelationpredictionsubgraph}, with 1,594 entities, 180
relations, and 4,245 triples. For comparison, ULTRA's pretraining corpus
(FB15k-237, WN18RR, CoDEx-Medium) contains roughly 130$\times$ more triples.

\textbf{Evaluation.} We evaluate on 40 inductive benchmarks under the
filtered ranking protocol and report absolute MRR: the 12 GraIL
inductive-(e) splits (FB15k-237, WN18RR, NELL-995, versions v1--v4), the
13 InGram inductive-(e,r) splits (FB, WK, NL, 25--100), and 15 extended
splits (ILPC 2022 small/large \citep{galkin2022ilpc}, HM 1k/3k/5k
\citep{hamaguchi2017knowledge, liu2021indigo}, FBNELL, Metafam, WikiTopics
MT1--4 \citep{zhou2023mtdea}). Every evaluation is zero-shot: the models are applied
as trained, with no fine-tuning and no target-side adaptation.

\textbf{Machine.} Every number in this paper, ours and the baseline's, was
produced on one NVIDIA A100 with 80 GB, one job at a time. The
30-minute training budget refers to that device: it is 1,800 s of
wall-clock training on a single A100, not a normalized compute unit.

\textbf{Baseline.} The baseline is ULTRA's released 3-graph
checkpoint,\footnote{\texttt{ultra\_3g.pth} from \texttt{https://github.com/DeepGraphLearning/ULTRA}.} reported in two columns. The \emph{re-run} column is that
checkpoint evaluated on the same machine with our evaluation code, on
the same splits, with the same filtered candidate sets and the same
tie-breaking as our models; it is the comparison we control end to end,
and the primary one. The \emph{published} column copies, split by split, the
zero-shot MRR of the same model from the appendix tables of the ULTRA
paper \citep{galkin2024foundationmodelsknowledgegraph}
(\texttt{results/ultra3g\_published.csv} records the source table of each
value). The two disagree: the published column is higher on 28 of the
40 benchmarks and by 0.008 MRR on average, with the largest gaps on
WN18RR v4 (0.611 published against 0.484 re-run), WikiTopics MT4:health
(0.624 against 0.557), NELL-995 v1 (0.785 against 0.727) and WN18RR v1
(0.648 against 0.593); the re-run is higher on Metafam (0.330 against
0.238) and on the three Hetionet-derived splits. The gap is not one we
can close. The published figures come from the original TorchDrug
implementation, while the released checkpoint is a PyTorch Geometric
re-implementation whose own README places it below the paper on the
inductive-(e) datasets (MRR 0.420 against 0.430 over 18 graphs) and at
parity on the inductive-(e,r) ones (0.344 against 0.345). Independent
re-runs land where ours does: the Gamma authors report ULTRA-3g at
0.5176 on the twelve GraIL splits under their own protocol
\citep{xin2025geometricstructuralknowledgegraph}, against our 0.5224 and the
published 0.5492. Every comparative table carries both columns; the
text quotes the re-run column, which is the controlled comparison. Read
against the published column instead, the comparison is less
favorable: the all-40 gap is 0.025 rather than 0.017 (Table 4), the GAT
trails on every family with more than one split except the Hetionet-derived
one (Table 3), and its lead
on the inductive-(e) benchmarks becomes parity (0.5457 against 0.5492,
one seed standard deviation). Only Section 5.2, which places the paper
among published KG foundation models, uses the published column in the
text.

\textbf{One excluded split.} The original suite has a 41st benchmark,
HM:indigo (458 relations), which we exclude on measured compute grounds:
evaluating a single model on it ran 4h38 without finishing, memory-bound
on the 80 GB A100. The decision was taken before any score of ours
existed on that split. The cut is not neutral: ULTRA is ahead of
comparable models there, so the exclusion shifts the aggregates in our
favor, by about +0.003 MRR (extended-split reference gap
$-$0.0807 $\rightarrow$ $-$0.0776; flat-aggregate reference gap $-$0.0388 $\rightarrow$ $-$0.0366). The
split remains in the published results, marked out of scope.

\textbf{Budget note.} The 30-minute cap binds only for the GAT, which
completes 14 of its 20 epochs; measured on the full evaluation, lifting
the cap to 20 epochs adds +0.004 MRR. The frugality of the recipe does
not hinge on the cap.

\textbf{Correction in this version.} The first version of this paper built
the test graph of eight of the 40 benchmarks with inverse edges counted
as facts. Our models were re-evaluated on these benchmarks from the same
frozen checkpoints, and the numbers of this version are the corrected
ones; Appendix B.9 describes the defect, lists what changed and states
the one exception. The ULTRA columns are unaffected.

\section{Results}

\subsection{Five textbook backbones, forty graphs}

\begin{table}[t]
\centering
\small
\caption{\textbf{filtered MRR, mean ± std over 3 seeds.} (From \texttt{results/backbone\_matrix.csv} / \texttt{broad\_eval\_matrix.csv}; full 15$\times$40 matrix in Appendix C. The two ULTRA rows are the re-run and published columns of Section 4; best value per column in bold.)}

\begin{tabular}{lrrrr}
\toprule
 & inductive-(e) (12) & inductive-(e,r) (13) & extended (15) & all 40 \\
\midrule
GAT & \textbf{0.5457 ± 0.0031} & 0.3316 ± 0.0019 & 0.2264 ± 0.0060 & 0.3564 ± 0.0037 \\
GINE (mean+max) & 0.5276 ± 0.0058 & 0.3161 ± 0.0053 & 0.1810 ± 0.0040 & 0.3288 ± 0.0047 \\
GINE (sum) & 0.5069 ± 0.0042 & 0.2980 ± 0.0044 & 0.1998 ± 0.0046 & 0.3238 ± 0.0039 \\
GraphSAGE & 0.4676 ± 0.0074 & 0.2832 ± 0.0036 & 0.1622 ± 0.0080 & 0.2932 ± 0.0040 \\
R-GCN & 0.4344 ± 0.0131 & 0.2071 ± 0.0144 & 0.1067 ± 0.0005 & 0.2376 ± 0.0083 \\
ULTRA (3-graph, re-run) & 0.5224 & 0.3446 & \textbf{0.2786} & 0.3732 \\
ULTRA (3-graph, published) & \textbf{0.5492} & \textbf{0.3517} & 0.2731 & \textbf{0.3815} \\
\bottomrule
\end{tabular}
\end{table}

Every backbone transfers: the weakest (R-GCN) still averages 0.2376
across 40 unseen graphs, far above chance on every family, and none
falls to random on any family, with one exception analyzed in
Section 5.3. The ordering (GAT \textgreater{} GINE-mean+max \textgreater{} GINE-sum \textgreater{} GraphSAGE \textgreater{}
R-GCN) is stable across evaluation panels, up to one swap: GINE-sum
overtakes GINE-mean+max on the extended panel.

The GAT is 0.0168 behind ULTRA on the full aggregate and ahead on the
twelve inductive-(e) benchmarks (0.5457 vs 0.5224). A 16-draw training
portfolio of the same GAT recipe averages 0.4301 ± 0.0048 on the
25-benchmark GraIL+InGram suite against 0.4299 for ULTRA; we summarize
this as \emph{matching} the dedicated foundation model in the single-small-KG
regime.

\subsection{Where the gap lives: families and regimes}

\begin{table}[t]
\centering
\small
\caption{\textbf{per-family filtered MRR, GAT (mean ± std) vs ULTRA, re-run and published columns.} Regime labels are relative to each model's training corpus: our models saw only FB15k-237-Inductive v1; ULTRA's corpus contains Freebase, WordNet, and Wikidata parents. Best value per row in bold.}

\begin{tabular}{lrrrr}
\toprule
family (n splits) & GAT & ULTRA re-run & ULTRA published & regime (ours / ULTRA) \\
\midrule
Freebase (8) & 0.4343 ± 0.0011 & 0.4361 & \textbf{0.4456} & in-family / in-family \\
WordNet (4) & 0.5545 ± 0.0036 & 0.5172 & \textbf{0.5745} & new KG / in-family \\
NELL (9) & 0.4688 ± 0.0047 & 0.4639 & \textbf{0.4764} & new KG / new KG \\
Wikidata (14) & 0.2689 ± 0.0098 & 0.2996 & \textbf{0.3039} & new KG / in-family \\
Hetionet-derived (3) & 0.0521 ± 0.0022 & \textbf{0.0659} & 0.0433 & new KG / new KG \\
Metafam (1) & 0.3165 ± 0.0458 & \textbf{0.3298} & 0.2380 & new KG / new KG \\
FBNELL (1) & 0.1061 ± 0.0003 & 0.4734 & \textbf{0.4850} & see Section 5.3 \\
\bottomrule
\end{tabular}
\end{table}

\begin{figure}[t]
\centering
\includegraphics[width=\textwidth]{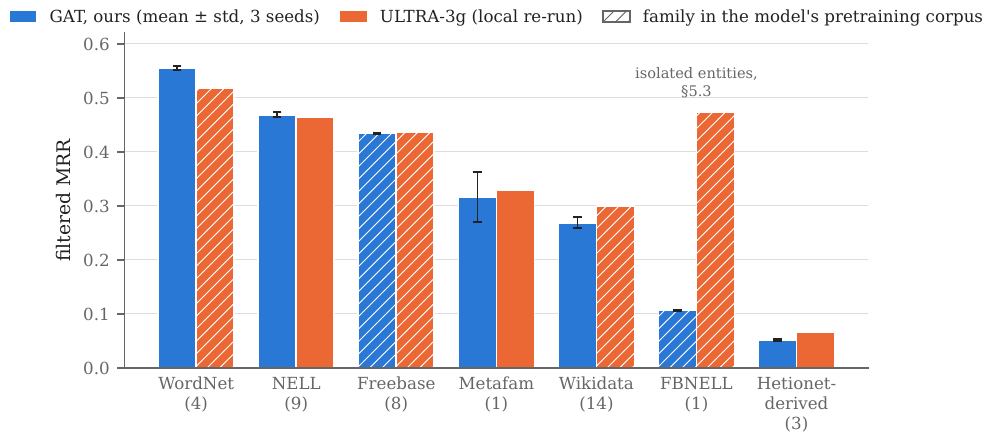}
\caption{\textbf{per-family filtered MRR, GAT (mean ± std over 3 seeds) versus the re-run ULTRA-3g, families ordered from our strongest to weakest.} Hatching marks a family present in the model's own pretraining corpus; FBNELL is the isolated-entity split of Section 5.3.}

\end{figure}

The GAT is ahead of ULTRA on WordNet and NELL, at parity on Freebase,
and behind on Wikidata, on the Hetionet-derived splits (hard for every
model; ULTRA itself is at 0.066), and on Metafam.

The all-40 aggregate is not a like-for-like comparison, and the
asymmetry runs against us. Our models saw one Freebase-derived graph;
ULTRA's corpus (FB15k-237, WN18RR, CoDEx-Medium) covers the Freebase,
WordNet, and Wikidata families. Of the 40 benchmarks, 27 belong to a
family ULTRA pretrained on, against 9 for us. Table 4 therefore reports
the aggregate at three nesting levels, using the regime labels assigned
when the evaluation matrix was built, before this comparison was
considered.

\begin{table}[t]
\centering
\small
\caption{\textbf{aggregates by how much of each family each model saw in pretraining.} Gap = GAT minus ULTRA, for each ULTRA column.}

\begin{tabular}{lrrrrrr}
\toprule
subset & n & GAT & ULTRA re-run & gap & ULTRA published & gap \\
\midrule
all benchmarks & 40 & 0.3564 ± 0.0037 & 0.3732 & $-$0.0168 & 0.3815 & $-$0.0251 \\
families unseen by us & 31 & 0.3444 ± 0.0046 & 0.3537 & $-$0.0094 & 0.3615 & $-$0.0172 \\
families unseen by both & 13 & 0.3609 ± 0.0062 & 0.3617 & $-$0.0008 & 0.3582 & +0.0028 \\
\bottomrule
\end{tabular}
\end{table}

The 13-benchmark row is the only symmetric comparison in the paper:
NELL (9 splits), the Hetionet-derived splits (3), and Metafam (1) are
new families for ULTRA exactly as they are for us. There the two models
are indistinguishable, the gap being well within our own seed-to-seed
spread. That number comes with two caveats: the subset is smaller and
dominated by the NELL family, so it measures a narrower question than
the all-40 row; and while the criterion is structural rather than
score-driven, the decision to report this level was taken after the
results were known. We therefore report all three rows and treat the
all-40 aggregate, where ULTRA leads by 0.017, as the primary
comparison.

Excluding the FBNELL split (analyzed next), the 39-split aggregate is
0.3628 vs 0.3706: the single pathological split accounts for more than
half of the overall gap to ULTRA.

\textbf{Where this sits among KG foundation models.} ULTRA is the reference
point of our evaluation, not the state of the art. Table 5 places the
GAT next to the published zero-shot figures of the four KGFMs that
followed ULTRA, on the 25 GraIL and InGram splits where all of them
report per-split numbers (their extended suites differ from ours). The
published rows are each paper's own figures, under its own evaluation
code and checkpoint-selection protocol, recomputed by us as unweighted
means over exactly these 12 and 13 splits; none of them is a
same-machine comparison, and the two ULTRA rows show how much that
alone can move a number. On the inductive-(e) splits the newer models
range from 0.008 (MOTIF) to 0.031 (Flock) above the published ULTRA; on
the inductive-(e,r) splits all five, ULTRA included, lie within 0.019 of
one another. Over the 25 splits the GAT, trained on one 4,245-triple
graph for 30 minutes, is 0.004 above the re-run ULTRA, 0.012 below the
published ULTRA, and 0.027 below TRIX, the strongest of the four.

\begin{table}[t]
\centering
\small
\caption{\textbf{zero-shot filtered MRR of published KG foundation models on the 25 GraIL and InGram splits, next to the GAT.} Published rows are each paper's own figures under its own evaluation code and checkpoint-selection protocol (Gamma selects its checkpoint on the target validation sets), recomputed as unweighted means over exactly these 12 and 13 splits (\texttt{results/kgfm\_published.csv} records the source table of each value); only the GAT and the re-run ULTRA are same-machine. All KGFMs pretrain on the same three graphs; KG-ICL, an in-context model pretrained on a different corpus, is omitted.}

\begin{tabular}{lrrrr}
\toprule
model & training data & inductive-(e) (12) & inductive-(e,r) (13) & all 25 \\
\midrule
GAT (ours, 3 seeds) & FB15k-237-Ind v1 & 0.5457 ± 0.0031 & 0.3316 ± 0.0019 & 0.4344 ± 0.0024 \\
ULTRA-3g, re-run & 3 KGs & 0.5224 & 0.3446 & 0.4299 \\
ULTRA-3g, published & 3 KGs & 0.5492 & 0.3517 & 0.4465 \\
TRIX, published & 3 KGs & 0.5790 & \textbf{0.3534} & \textbf{0.4617} \\
MOTIF, published & 3 KGs & 0.5567 & 0.3346 & 0.4412 \\
Gamma, published & 3 KGs & 0.5581 & 0.3527 & 0.4513 \\
Flock, published & 3 KGs & \textbf{0.5797} & 0.3435 & 0.4569 \\
\bottomrule
\end{tabular}
\end{table}

\subsection{The one failure mode: isolated entities}

On FBNELL, all fifteen models collapse to MRR 0.076--0.106 with Hits@1
exactly 0, the only anomaly in the 600-cell matrix. FBNELL is the only
split of the 40 whose test inference graph contains isolated entities:
exactly five, each appearing in no triple of any split and in exactly
one test triple (five queries whose answer is isolated, one whose source
is). Our training graph contains no such
entity (each of its 1,594 entities carries at least one fact), and
neither does any other inference graph of the 40. Under seeded
propagation no message ever reaches an isolated entity, so its state
remains at its initialization, identical for all five (entities are
anonymous) and constant across the 1,194 queries of the split. Under the
plain readout that constant state happens to outscore every connected
candidate, so the five occupy ranks 1--5 of every query: the rank
distributions show no query at ranks 1--4 and a mode at rank 6 (454--470 of
1,194 queries across the three GAT seeds, which is rank 1 once the
preempting block is discounted). Five entities without a single observed
fact cost the backbones between 0.23 and 0.39 MRR (0.38 for the GAT,
estimated by discounting the five preempting ranks). We report the 0.106 as measured, since removing the isolated
entities from the candidate pool would edit the protocol in our favor on
the one split where it hurts us.

ULTRA is immune, and the contrast explains why. What matters is that the
degenerate state be scored \emph{below} the true answers, and that property
must be learned. ULTRA's readout conditions on the query, and the state
it must downrank is not rare in its regime: for a source-conditioned
model, every candidate beyond the reach of the propagation carries no
query signal, on any graph, so a model that scored that state high would
collapse everywhere; ULTRA's training keeps that state out of the top
rather than at the bottom. For
our models the degenerate state requires a degree-0 entity, which no
training graph we used contains, so the score a trained model assigns to
it is arbitrary: the same recipe trained on CoDEx-Small (Section 5.4)
happens to show no preemption on FBNELL (0.357 versus our 0.106)
although CoDEx-Small contains no low-degree entity either, an arbitrary
sign rather than a fix. What no model can
learn is the handful of queries whose true answer \emph{is} one of the five:
the five candidates are structurally identical, so any
permutation-equivariant model scores them identically and those queries
are decided by tie-breaking, for ULTRA as for us, as ULTRA-3g's own
ranks confirm: it scores the five identically and places them
mid-ranking, so the queries whose answer is isolated land at ranks 12 to
530 (Appendix C.4). ULTRA's advantage on the split is earned on the other
queries, where isolated candidates merely need to stay out of the way. Our models condition at the
propagation level only (Section 3.2), which cannot reach a disconnected
candidate by construction. Six models of the same family do have a
query-conditioned readout: the generic model of the probe, in its three
pretraining draws, and three models trained on our single training
graph with that model's recipe (Appendix D.3). None was trained on a
graph containing an isolated entity, and none of them shows the
preemption: on FBNELL all six place their mode at rank 1 and score
between 0.31 and 0.43 MRR with Hits@1 between 0.24 and 0.34, against
0.08 to 0.11 and Hits@1 of exactly zero for the fifteen plain-readout
models (Table B2). The test is not a pure ablation, since these models
also differ from those of Table 2 by their training recipe, so we read
it as a confirmation of the mechanism rather than as a measurement of
the readout alone. One trace remains: for the three single-graph models,
the five queries whose answer is isolated still land at ranks 5 to 20, so
the isolated block stays grouped just below the top on exactly those
queries, while the three-graph models spread them between ranks 28 and
282 (Table B2).

\begin{figure}[t]
\centering
\includegraphics[width=0.62\textwidth]{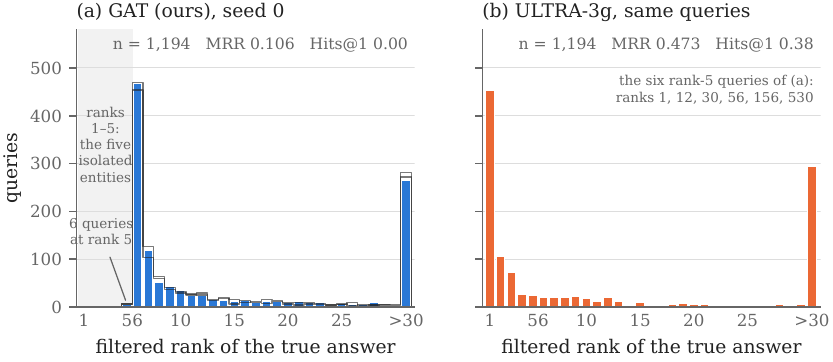}
\caption{\textbf{FBNELL, distribution of the filtered rank of the true answer over the 1,194 queries. (a) GAT seed 0 (bars) with seeds 1 and 2 as outlines; (b) ULTRA-3g on the same queries, same filter.} Panel (a): no query lands at ranks 1--4: the five isolated entities occupy them; the six queries at rank 5 are the five whose answer is isolated and the one whose source is; the mode is rank 6. Panel (b): ULTRA-3g (MRR 0.473, Hits@1 0.38) has its mode at rank 1 and no gap; the six rank-5 queries of panel (a) land at ranks 1, 12, 30, 56, 156 and 530.}

\end{figure}

\subsection{Does the choice of training graph matter?}

The models of Section 5 train on one Freebase-derived graph, and Table 3
locates their weakness on the Wikidata and Hetionet-derived families.
This section measures how much of that profile is due to the training
graph itself: we retrain the same recipe on other single training graphs
and read the result per evaluation family (full per-family table in
Appendix C).

The informative candidate is CoDEx-Small, a Wikidata-family graph,
retrained with 3 seeds and gradient-step-matched to the reference. If
family proximity between training and evaluation graphs drove transfer,
this model should gain exactly where ours is weak; instead it loses
0.003 on the Wikidata-family benchmarks and 0.059 on the 40-graph mean.
Two larger candidates, ConceptNet100k and AristoV4, do not fit
full-graph training in 80 GB and were compared in a sampled-subgraph
regime against a reference retrained the same way; neither helps on any
weak family, and since their numbers are not comparable to the rest of
the paper they stay in the appendix.

The choice of training graph therefore explains neither the successes
nor the remaining gap: the transfer of Table 2 does not come from a
fortunate match between training and evaluation families, since the
in-family candidate gains nothing even on its own family, and the gap
to ULTRA is not closed by a better-chosen single graph. ULTRA's own
pretraining study points the same way from the other side: varying the
pretraining mix from one to eight graphs, its authors observe
performance saturating beyond three and leave a more principled choice
of mix to future work \citep{galkin2024foundationmodelsknowledgegraph}.
Training mixtures on the reified representation are the natural next
experiment (Section 7).

\section{The same vocabulary reads a relational database}

Section 3.1 maps a relational database into the reified vocabulary: rows
become entities, foreign-key columns relation types, foreign-key cells
facts, and tables entity-type nodes. The models available for the probe
are pretrained on knowledge graphs only, whose vocabulary has no entity
type, so they cannot read a \texttt{CLASS} node or an \texttt{is\_a} edge: the probe
therefore feeds them the database in its knowledge-graph form (rows,
foreign-key columns, foreign-key cells), and uses the tables only to
constrain the candidate set (Appendix D.1). To test whether this mapping
yields
a shared vocabulary rather than merely a shared notation, we take a
model pretrained on knowledge graphs only, reify an unseen database, and
rank foreign-key targets with no features, no fine-tuning, no schema
text and no in-context labels.

The regime is deliberately impoverished. Cell values are where most of
a database's information sits, and
schema text and in-context labels are the two mechanisms through which
relational foundation models actually transfer (Section 2, Table 1). A
model given none of the three has only the key structure left. Existing
relational foundation models cannot run here at all: strip the text and
the labeled context and they have no input. Any signal measured in this
regime is therefore attributable to structure alone.

We summarize the probe here; Appendix D gives the protocol and the
per-key results on both databases. The transferring model is a generic
corpus-pretrained backbone of this family, not one of the models of
Section 5. On two unseen databases (243k and 10.06M reified nodes), it
ranks foreign-key targets far above both controls on entity-referencing
keys: Hits@10 between 0.78 and 1.00 on the four driver and constructor keys
of the qualifying and results tables of a Formula-1 database, against 0.00 for the same architecture with
random initialization, and well above a degree heuristic. The result
holds across pretraining seeds and on both databases, and the failures
are consistent: the keys whose targets form a large class of
near-interchangeable events (races, posts) stay near random, as nothing
in the bare key structure distinguishes one race from another.

Nothing in the setup is exclusive to our models: the foreign-key graph
of Step 1 (Section 3.1) is an ordinary multi-relational graph that a KG
foundation model can consume directly. We therefore ran the three
released ULTRA checkpoints zero-shot on the raw foreign-key graph, with
the same queries, candidates and controls (Appendix D). The checkpoint
pretrained on the same three graphs as our model (ULTRA-3g) never rises
above the degree heuristic (1 of 13 keys on rel-f1, 0 of 11 on
rel-stack). The more broadly pretrained checkpoints do extract
structure: ULTRA-50g rises above both controls on 5 keys of each
database, substantially (Hits@10 above 0.10) on 3 of rel-f1 and 4 of
rel-stack, with a profile different from our model's. Our
generic model stays ahead on three of the four driver and constructor
keys of rel-f1 (Hits@10 gaps of 0.20 to 0.43, non-overlapping bootstrap
CIs; parity on the fourth, 1.00 versus 0.95) and on the three author
keys of rel-stack (gaps of 0.38 to 0.82, non-overlapping CIs), ULTRA-50g
is ahead on user votes and parent posts (0.73 vs 0.32, 0.27 vs 0.00),
and the two are at parity on the post keys of comments and edit history.
The two representations thus extract partially disjoint signals from the
same graph, and ULTRA's zero-shot behavior is strongly
checkpoint-dependent on individual keys (0.13, 0.02 and 0.83 on
posts.OwnerUserId across the three checkpoints, against 0.83 for our
model). Finally, the models of Section 5, pretrained on FB15k-237 v1
alone, do not transfer at all (Hits@10 0.00 on every entity key, three
seeds): the probe signal is a property of the diverse pretraining corpus
as much as of the representation.

The corpus dependence can be isolated. With the same backbone,
recipe and reified representation, models pretrained on FB15k-237 v1 alone
transfer on none of the entity-referencing rel-f1 keys (six models: two
recipes, three seeds each) and on a recipe- and seed-dependent subset of the rel-stack
keys (two post keys reliably with the Section 5 recipe; three author keys
for one seed in three with the corpus recipe). Pretrained on three
transductive KGs, the same backbone transfers on all ten strong keys of
both databases for each of three pretraining seeds. The multi-graph corpus
is what makes the transfer complete and reliable; the reified
representation is what lets a three-graph model extract it where ULTRA-3g,
pretrained on the same three graphs, does not.

The result remains a probe, run on two databases, with one reified model
family and one external KG foundation model, and no comparison against a
supervised specialist. We will pursue these experiments in future work,
shifting the emphasis to what the shared vocabulary uniquely provides: a
single model that reads both data models natively, pretrained on
knowledge graphs and relational databases jointly and applied to either,
rather than a setup in which any KG foundation model can stand in for
ours.

\section{Limitations}

\textbf{The gap that remains.} ULTRA is ahead by 0.017 overall, concentrated
on the Wikidata and Hetionet-derived families (Table 3), and Section 5.4
shows no single alternative training graph closes it; the KGFMs that
followed ULTRA publish figures a further 0.008 to 0.031 above ULTRA's on
the inductive-(e) splits (Table 5). Vanilla transfer
from one small KG is demonstrated in-distribution of the
Freebase/WordNet/NELL families; ULTRA's multi-graph invariants
generalize further. Multi-graph training mixtures on the reified
representation are the natural next experiment and are out of scope
here.

\textbf{Isolated entities.} Our deliberately minimal models omit
query-conditioned readout, which Section 5.3 identifies as what protects
ULTRA from degree-0 entities and confirms on six models of the family
that have one. We stop short of calling it necessary: one plain-readout
model escaped the pathology through its training graph alone, with a
sign nothing guarantees. What remains open is the residual gap to ULTRA
on that split (0.43 against 0.47 for the best of the six) and the
grouping of the isolated candidates that persists, for the single-graph
models, on the queries whose own answer is isolated; a controlled
ablation that changes the readout and nothing else would separate the
readout from the recipe.

\textbf{Cost accounting.} The 30-minute budget is a property of the training
graph's size as much as of the method (Section 5.4); the frugality holds
in the demonstrated regime and we do not present it as a scaling law. Reification itself is
not free: it multiplies node count by roughly $\times$20 and doubles the graph
diameter, so a reified graph needs twice the depth of the original for
the same reach, and full-graph propagation stops fitting in memory well
before a plain KG would. Section 5.4 hits exactly that wall on the
100k-fact training candidates.

\textbf{Scope.} All results are structural link prediction on featureless
graphs; the database probe (Section 6) is evidence for the vocabulary,
not for a general capability on relational databases. It also leaves the
entity-type nodes of Section 3.1 untested: exercising them requires a
model pretrained on data that has tables, the joint KG-and-database
pretraining announced as future work.

\section{Conclusion}

We have shown that zero-shot inductive link prediction across knowledge
graphs does not require a dedicated foundation-model architecture; it
requires a representation in which datasets share a vocabulary.
Reification provides one, with three node kinds and six meta-relations
(four and eight for databases), and on it five textbook GNNs,
trained on a single 4,245-triple knowledge graph for 30 minutes on one
NVIDIA A100, transfer to 40 inductive benchmarks. The best of them, an
off-the-shelf GAT, matches ULTRA where the comparison is symmetric
(0.3609 versus 0.3617 on the 13 benchmarks whose families neither model
saw in pretraining) and trails it by 0.017 on the full 40-benchmark
aggregate, more than half of that gap tracing to a single split with
isolated entities. The same
vocabulary reads a relational database, and a preliminary probe on two
unseen databases shows a model of this family pretrained on three
knowledge graphs ranking foreign-key targets far above
random-initialization and degree controls. The
transfer mechanism, once moved from the architecture into the data, is
available to any node-inductive GNN. We release the code and the
evaluation pipeline for all 40 benchmarks at
\url{https://github.com/getorbital/REIFM}, and the checkpoints (the 15 KG
models of Table 2 and the six models of the probe) at
\url{https://huggingface.co/itmatr/REIFM}.

\begin{center}\rule{0.5\linewidth}{0.5pt}\end{center}

\appendix
\counterwithin{table}{section}
\renewcommand{\thetable}{\thesection\arabic{table}}
\section{Protocol details}

\textbf{Frozen recipe.} Dimension 64, 12 layers, dropout 0.2, cosine learning
rate schedule with base rate $5\cdot 10^{-4}$, weight decay 0.01, batch size 16,
20 epochs capped at 1,800 s, 1,000 validation queries, full-graph
propagation, plain readout. Training graph: the training split of
FB15k-237-Inductive v1. Only \texttt{backbone} and \texttt{seed} vary across the 15
models. Model selection is Tier-1: the checkpoint maximizing validation
MRR on the training graph's own validation split, never on any test
split. Seeds \{0, 1, 2\} were fixed a priori for every backbone; no seed
was rejected, re-drawn, or replaced, including the weakest draw of each
family. Table A1 lists the argument block as stored with each
checkpoint; the released checkpoints ship with the same file.

\begin{table}[p]
\centering
\small
\caption{frozen argument block of the 15 KG models (from the stored \texttt{results.json}; only \texttt{backbone} and \texttt{seed} vary). Two later-introduced flags, \texttt{class\_node} and \texttt{ent\_features\_dir}, are absent from these runs, which is their default (off / none).}
\label{tab:appA_recipe}
\begin{tabular}{lll}
\toprule
argument & value & note \\
\midrule
\texttt{backbone} & gat & varies: gat · ginemax · gine\_sum · sage · rgcn \\
\texttt{seed} & 0 & varies: 0 · 1 · 2 \\
\texttt{dim} & 64 &  \\
\texttt{layers} & 12 &  \\
\texttt{agg} & sum & used by the PyG GINE (sum) and GraphSAGE convolutions only \\
\texttt{dropout} & 0.2 &  \\
\texttt{edge\_dropout} & 0.0 & off \\
\texttt{lr} & 0.0005 &  \\
\texttt{cosine} & True &  \\
\texttt{wd} & 0.01 &  \\
\texttt{bs} & 16 &  \\
\texttt{accum\_steps} & 1 &  \\
\texttt{grad\_clip} & 0.0 & off \\
\texttt{epochs} & 20 &  \\
\texttt{max\_train\_secs} & 1800.0 & binding for GAT only (14 epochs) \\
\texttt{train} & FB15k237Inductive:v1 &  \\
\texttt{select\_eval} & FB15k237Inductive:v1 & Tier-1 \\
\texttt{valid\_every} & 1 &  \\
\texttt{valid\_queries} & 1000 &  \\
\texttt{pos\_mask} & True &  \\
\texttt{sampled} & False & off: full-graph propagation \\
\texttt{qc\_readout} & False & off: plain readout \\
\texttt{qc\_complex} & False & off \\
\texttt{rel\_graph\_layers} & -1 & $-$1 = off \\
\texttt{use\_cofact} & False & off \\
\texttt{rel\_pe} & 0 & off \\
\texttt{ent\_pe} & 0 & off \\
\texttt{jk} & False & off \\
\texttt{swa} & False & off \\
\texttt{nre\_p} & 0.0 & off \\
\texttt{head\_margin\_k} & 0 & off \\
\texttt{grad\_ckpt} & False & off \\
\bottomrule
\end{tabular}
\end{table}

\textbf{Machine.} One NVIDIA A100 (80 GB), one job at a time, for every
number in this paper including the ULTRA re-run (Section 4). Timings
elsewhere in the paper --- the 1,800 s training cap, the 4h38 measured on
the excluded split, the per-gradient-step costs of Appendix B --- all
refer to that device.

\textbf{Code non-drift control.} Eleven of the fifteen checkpoints were
trained three weeks before the remaining four, and four commits had
touched the training and model paths in between, each claiming to be a
no-op under our flag settings. Rather than trust that, we verified it in
two steps before training the missing seeds, with the acceptance
threshold (0.005 MRR) fixed in advance.

\emph{Evaluation path.} Re-evaluating an existing checkpoint on five splits
with the current code reproduced its stored results to within 0.0009
(FB15k-237-v1 0.4951 $\rightarrow$ 0.4954; WN18RR-v1 0.6605 $\rightarrow$ 0.6605; NELL-v1
0.6182 $\rightarrow$ 0.6180; FB-InGram-25 0.3021 $\rightarrow$ 0.3021; NL-InGram-0 0.3800 $\rightarrow$
0.3791).

\emph{Training path.} Retraining one configuration (R-GCN, seed 0) bit-for-bit
under the current code reproduced its best validation MRR to within
0.0003 (0.4618 at epoch 16 against 0.4621 at epoch 18), with identical
parameter count, the same number of epochs, and superposed loss
trajectories. R-GCN was chosen because none of the paper's main results
rests on it. The
replica is archived as a reproducibility control and does \emph{not} enter
any table; the original checkpoint stays in the matrix, so no selection
is made after the fact.

Conformance of all fifteen checkpoints to the frozen recipe was then
checked field by field. One audit subtlety is worth recording: two
arguments introduced by later commits do not exist in the older runs'
stored configurations, and their absence is the default the protocol
requires. A naive comparison that treats ``key absent'' as a difference
reports eleven spurious deviations.

\section{Transparency notes}

\textbf{B.1 The excluded split.} Section 4 gives the decision and its two
deltas. For completeness, the 41-benchmark aggregates that the exclusion
removes are: ULTRA 0.3747 $\rightarrow$ 0.3732 and the internal reference model
0.3359 $\rightarrow$ 0.3366 on the flat mean (gap $-$0.0388 $\rightarrow$ $-$0.0366); on the
extended panel, the reference-to-ULTRA gap moves from $-$0.0807 over 16
splits to $-$0.0776 over 15. HM:indigo is the fourth most unfavorable
extended split for that reference ($-$0.1270); the most unfavorable one,
WikiTopics-MT3:infra ($-$0.1895), stays in scope. No model of this paper
was ever evaluated on HM:indigo; it appears in Table C1 with ULTRA's
score only.

\textbf{B.2 FBNELL, as measured.} Table B1 lists the fifteen FBNELL cells
exactly as evaluated under the protocol of Section 4. Removing the five
isolated entities from the candidate pool would raise every one of these
cells by 0.23 to 0.39 MRR (Section 5.3); we did not do it, because it
would edit the protocol in our favor on the one split where it hurts us.

\begin{table}[htbp]
\centering
\scriptsize
\caption{the fifteen FBNELL cells as measured (1,194 queries; the five isolated entities occupy ranks 1–5 of every query).}
\label{tab:appB_fbnell_cells}
\adjustbox{max width=\textwidth}{%
\begin{tabular}{lrrrrr}
\toprule
backbone & seed & MRR & Hits@1 & Hits@3 & Hits@10 \\
\midrule
GAT & 0 & 0.1063 & 0.0000 & 0.0000 & 0.6030 \\
GAT & 1 & 0.1057 & 0.0000 & 0.0000 & 0.5930 \\
GAT & 2 & 0.1062 & 0.0000 & 0.0000 & 0.6022 \\
GINE (mean+max) & 0 & 0.1024 & 0.0000 & 0.0000 & 0.5737 \\
GINE (mean+max) & 1 & 0.1015 & 0.0000 & 0.0000 & 0.5687 \\
GINE (mean+max) & 2 & 0.1034 & 0.0000 & 0.0000 & 0.5729 \\
GINE (sum) & 0 & 0.0996 & 0.0000 & 0.0000 & 0.5704 \\
GINE (sum) & 1 & 0.0981 & 0.0000 & 0.0000 & 0.5570 \\
GINE (sum) & 2 & 0.0971 & 0.0000 & 0.0000 & 0.5528 \\
GraphSAGE & 0 & 0.0958 & 0.0000 & 0.0000 & 0.5410 \\
GraphSAGE & 1 & 0.0958 & 0.0000 & 0.0000 & 0.5436 \\
GraphSAGE & 2 & 0.0961 & 0.0000 & 0.0000 & 0.5377 \\
R-GCN & 0 & 0.0763 & 0.0000 & 0.0000 & 0.3928 \\
R-GCN & 1 & 0.0829 & 0.0000 & 0.0000 & 0.4146 \\
R-GCN & 2 & 0.0781 & 0.0000 & 0.0000 & 0.3911 \\
\bottomrule
\end{tabular}
}
\end{table}

Table B2 gives, under the same protocol and on the same machine, the six
models with a query-conditioned readout evaluated for Section 5.3
(\texttt{results/fbnell\_qc\_readout.csv}); the candidate pool is untouched here
as well. The last column lists the filtered ranks of the six queries
whose answer (five) or source (one) is an isolated entity, in the order
of the test file; the third listed is the query whose source is
isolated.

\begin{sidewaystable}
\centering
\scriptsize
\caption{the six query-conditioned-readout models of Section 5.3 on FBNELL, as measured (1,194 queries, same protocol and machine as Table B1; the five isolated entities stay in the candidate pool). 3-KG generic = the probe's generic model, pretraining draws A, B, C (Appendix D.3); 1-KG, corpus recipe = the same backbone and recipe trained on FB15k-237-Inductive v1 alone.}
\label{tab:appB2_fbnell_qc}
\adjustbox{max width=0.92\textheight, max totalheight=0.9\textwidth}{%
\begin{tabular}{llrrrrrrl}
\toprule
model & seed & MRR & Hits@1 & Hits@3 & Hits@10 & queries at ranks 1–4 & mode rank & ranks of the 6 isolated queries \\
\midrule
3-KG generic & A & 0.4296 & 0.3317 & 0.4732 & 0.6164 & 609 & 1 & 155, 59, 1, 282, 33, 116 \\
3-KG generic & B & 0.4259 & 0.3308 & 0.4690 & 0.6114 & 592 & 1 & 187, 172, 1, 277, 36, 247 \\
3-KG generic & C & 0.4332 & 0.3442 & 0.4682 & 0.6039 & 597 & 1 & 161, 69, 1, 212, 28, 84 \\
1-KG, corpus recipe & 0 & 0.3105 & 0.2353 & 0.2898 & 0.5343 & 353 & 1 & 5, 5, 1, 5, 6, 5 \\
1-KG, corpus recipe & 1 & 0.3791 & 0.3065 & 0.3735 & 0.5628 & 459 & 1 & 5, 7, 5, 7, 6, 6 \\
1-KG, corpus recipe & 2 & 0.3882 & 0.3090 & 0.4045 & 0.5561 & 510 & 1 & 5, 11, 5, 20, 12, 6 \\
\bottomrule
\end{tabular}
}
\end{sidewaystable}

\textbf{B.3 Provenance of the evaluation cells.} Of the 600 cells of the
main matrix, the 75 GAT cells on the 25 GraIL and InGram benchmarks are
the frozen evaluations of the same checkpoints from the earlier campaign
(same harness, same machine); the evaluation-path control of Appendix A
bounds the drift between the two dates at 0.0009. On the two smallest
splits (NELL-995-v1, 402 queries; NL-InGram-0, 1,526) GPU evaluation is
not deterministic: re-running the same checkpoint with the same code
changes the rank of 30--40 (resp. about 150) queries and moves the MRR by
up to ±0.006 (resp. ±0.002); the released integrity check therefore uses
the pre-registered 0.005 threshold, and reproduces 44 of its 75 cells
exactly. The remaining 525
cells were evaluated in one campaign. The re-run ULTRA column was
evaluated once on the same machine and never re-run. One extended cell
(GAT seed 2, ILPC2022-small) had also been evaluated independently three
days earlier and reproduced within 0.001, before the correction of
Appendix B.9.

\textbf{B.4 Gradient-step matching for CoDEx-Small (Section 5.4).} Under the
frozen recipe, one gradient step on CoDEx-Small costs 13$\times$ one step on
the reference graph (3.52 s versus 0.27 s per batch, measured on the
A100), so the 1,800 s cap would allow 511 steps against 7,434 for the
reference, 7 \% of its training. We therefore matched steps rather than
seconds: 14 epochs of 531 batches, the reference's exact schedule, with
the same 14 selection points, about 7.3 h per seed. Gradient
checkpointing was enabled to fit the 35,000-node reified graph in memory
(24 GB instead of an out-of-memory failure); it changes memory use, not
the computation. Everything else in the recipe is unchanged, and
selection uses CoDEx-Small's own validation split.

\textbf{B.5 The sampled regime (Section 5.4).} ConceptNet100k and AristoV4
do not fit full-graph training in 80 GB. Both were trained with
sampled-subgraph propagation, and compared against the reference recipe
retrained in that same regime on FB15k-237-Inductive v1 (one seed each).
Bounded neighborhoods cap what any model can reach: the sampled
reference averages 0.2774 over the 40 benchmarks against 0.3564 for the
full-graph GAT. The two sampled columns of Table C3 are therefore
comparable to their own reference only, never to Table 2.

\textbf{B.6 The pretraining draws behind the probe.} The generic model of
Section 6 was pretrained four times with the same recipe (Appendix D.3):
the original run and three re-pretrainings with seeds 0, 1 and 2. A
sanity gate, fixed before any probe was run, required a re-pretraining
to reach 95 \% of the original run's selection score (0.4310 on the mean
validation MRR of the three selection splits, hence a threshold of
0.4095). Seeds 0 and 2 passed (0.4337 and 0.4410); seed 1 did not
(0.3788 after its full 6 h budget), and by the pre-registered rule no
probe result was drawn from it. The three passing models are the three
seeds of Table D6; the failed draw is reported here rather than averaged
away, as a data point on the seed variance of the corpus recipe.

\textbf{B.7 Reporting rules.} All aggregates are unweighted means of
per-benchmark filtered MRR; standard deviations are over the three
seeds, computed per aggregate (mean over benchmarks per seed, then
standard deviation across seeds, n $-$ 1 denominator). Every ULTRA figure
labeled \emph{re-run} is the local evaluation of the released checkpoint
(Section 4); every figure labeled \emph{published} is copied, split by split,
from the appendix tables of the ULTRA paper, with the source table of
each value recorded in \texttt{paper\_short/results/ultra3g\_published.csv}; the
KGFM rows of Table 5 are copied the same way from the respective papers
(\texttt{paper\_short/results/kgfm\_published.csv}). Regime labels (Tables 3, 4 and C1)
were assigned when the evaluation matrix was assembled, before the
aggregates of Table 4 were considered.

\textbf{B.8 Declaration on generative AI.} Anthropic's Claude, through Claude
Code, was used throughout this work as a coding and writing assistant.
It wrote most of the experiment code, ran and logged the experiments
under protocols fixed in advance by the author, and drafted and edited
the text, tables and figures. The research questions, the design of the
representation, the evaluation protocol, every decision on what to run
and what to claim, and the final text were set, reviewed and approved by
the author, who takes full responsibility for the content.

\textbf{B.9 Correction to the test graphs of eight benchmarks.} The first
version of this paper (arXiv v1) built the test graph of eight benchmarks
from an edge list that still contained inverse edges. Our loader removed
them using the number of relations of the dataset, whereas the inference
graph numbers its inverse relations from its own count, which is smaller on
these benchmarks: WKIngram:100, FBIngram:100, ILPC2022-small, WikiTopics
MT1:tax, NLIngram:0, NLIngram:100, ILPC2022-large and WikiTopics MT4:health.
Part of their inverse edges entered the graph as facts, from twice the
number of facts on WKIngram:100 to 49 extra facts on MT4:health. No test
triple was involved and the training graphs were unaffected, so no model
was retrained. The 15 models of Table 2 and the three CoDEx-Small models
of Section 5.4 were re-evaluated on the eight benchmarks from the same
frozen checkpoints with the corrected loader; this changes Tables 2 to 4,
the GAT row of Table 5, Figure 2, Tables C1 and C2, and the CoDEx-Small
column of Table C3. The ULTRA columns and the other rows of Table 5 use
the loaders of their own code and were not affected; the FBNELL cells
(Section 5.3, Tables B1 and B2) and the database probe (Section 6,
Appendix D) are unchanged. Individual cells move by up to 0.081 (R-GCN,
seed 2, on FBIngram:100); the 40-benchmark mean of the GAT
moves from 0.3565 ± 0.0049 to 0.3564 ± 0.0037, its distance to the re-run
ULTRA from 0.0167 to 0.0168, and the 13-benchmark mean of Table 4 from
0.3600 to 0.3609. The other models evaluated by our code were
re-evaluated the same way: the thirteen further draws of the portfolio of
Section 5.1 (mean 0.4293 $\rightarrow$ 0.4301), the three models of the sampled regime
(Table C3, Appendix B.5) and the 20-epoch GAT trainings behind the budget
note of Section 4 (+0.003 $\rightarrow$ +0.004). These last were re-evaluated on seven
of the eight benchmarks and not on ILPC2022-large, where the cells of the
15 models of Table 2 move by at most 0.004. The cells of the first version
(\texttt{results/v1/}) are released with the code.

\section{Full result matrices}

\textbf{C.1 Per-benchmark results.} Table C1 gives, for each of the 40
benchmarks, the mean and standard deviation over the three seeds of
each backbone, the locally re-run ULTRA-3g, and the regime labels used
in Table 4. Table C2 gives the underlying 15 $\times$ 40 matrix, one cell per
(backbone, seed, benchmark). Both are derived from the released
\texttt{broad\_eval\_matrix.csv}, which also carries Hits@1/3/10 per cell.

\begin{sidewaystable}
\centering
\scriptsize
\caption{filtered MRR on the 40 benchmarks, mean $\pm$ std over the 3 seeds of each backbone, and ULTRA-3g in its two columns (Section 4): re-run locally from the released checkpoint, and as published in the ULTRA paper's appendix tables. Regime labels: train = the training graph's own split; in-family = a family present in the model's pretraining corpus; new KG = family unseen; partial = FBNELL (Freebase and NELL). Panel order: 12 GraIL, 13 InGram, 15 extended.}
\label{tab:appC_per_split}
\adjustbox{max width=0.92\textheight, max totalheight=0.9\textwidth}{%
\begin{tabular}{lllrrrrrrr}
\toprule
split & family & regime ours / ULTRA & GAT & GINE (mean+max) & GINE (sum) & GraphSAGE & R-GCN & ULTRA-3g re-run & ULTRA-3g published \\
\midrule
FB15k237Inductive:v1 & Freebase & train / in-family & 0.5257 $\pm$ 0.0062 & 0.4979 $\pm$ 0.0053 & 0.5074 $\pm$ 0.0085 & 0.4982 $\pm$ 0.0070 & 0.4581 $\pm$ 0.0025 & 0.4863 & 0.498 \\
FB15k237Inductive:v2 & Freebase & in-family / in-family & 0.5322 $\pm$ 0.0023 & 0.5159 $\pm$ 0.0053 & 0.5034 $\pm$ 0.0109 & 0.4867 $\pm$ 0.0023 & 0.4458 $\pm$ 0.0064 & 0.5010 & 0.512 \\
FB15k237Inductive:v3 & Freebase & in-family / in-family & 0.5052 $\pm$ 0.0010 & 0.4786 $\pm$ 0.0028 & 0.4730 $\pm$ 0.0052 & 0.4571 $\pm$ 0.0005 & 0.4224 $\pm$ 0.0085 & 0.4821 & 0.491 \\
FB15k237Inductive:v4 & Freebase & in-family / in-family & 0.4836 $\pm$ 0.0017 & 0.4623 $\pm$ 0.0026 & 0.4502 $\pm$ 0.0075 & 0.4352 $\pm$ 0.0034 & 0.3823 $\pm$ 0.0031 & 0.4770 & 0.486 \\
NELLInductive:v1 & NELL & new KG / new KG & 0.6262 $\pm$ 0.0137 & 0.6219 $\pm$ 0.0112 & 0.5778 $\pm$ 0.0203 & 0.7806 $\pm$ 0.0231 & 0.4674 $\pm$ 0.1203 & 0.7270 & 0.785 \\
NELLInductive:v2 & NELL & new KG / new KG & 0.5595 $\pm$ 0.0071 & 0.5175 $\pm$ 0.0072 & 0.4803 $\pm$ 0.0051 & 0.3236 $\pm$ 0.0195 & 0.3544 $\pm$ 0.0155 & 0.5247 & 0.526 \\
NELLInductive:v3 & NELL & new KG / new KG & 0.5624 $\pm$ 0.0051 & 0.5366 $\pm$ 0.0072 & 0.4980 $\pm$ 0.0058 & 0.4013 $\pm$ 0.0127 & 0.3427 $\pm$ 0.0158 & 0.5112 & 0.515 \\
NELLInductive:v4 & NELL & new KG / new KG & 0.5356 $\pm$ 0.0033 & 0.5028 $\pm$ 0.0119 & 0.4662 $\pm$ 0.0102 & 0.2647 $\pm$ 0.0164 & 0.3182 $\pm$ 0.0249 & 0.4905 & 0.479 \\
WN18RRInductive:v1 & WordNet & new KG / in-family & 0.6567 $\pm$ 0.0065 & 0.6558 $\pm$ 0.0070 & 0.6304 $\pm$ 0.0224 & 0.5656 $\pm$ 0.0419 & 0.6009 $\pm$ 0.0134 & 0.5929 & 0.648 \\
WN18RRInductive:v2 & WordNet & new KG / in-family & 0.6335 $\pm$ 0.0054 & 0.6340 $\pm$ 0.0098 & 0.6127 $\pm$ 0.0134 & 0.5765 $\pm$ 0.0173 & 0.5984 $\pm$ 0.0104 & 0.6205 & 0.663 \\
WN18RRInductive:v3 & WordNet & new KG / in-family & 0.3304 $\pm$ 0.0009 & 0.3106 $\pm$ 0.0086 & 0.3097 $\pm$ 0.0013 & 0.3095 $\pm$ 0.0019 & 0.2771 $\pm$ 0.0064 & 0.3712 & 0.376 \\
WN18RRInductive:v4 & WordNet & new KG / in-family & 0.5974 $\pm$ 0.0034 & 0.5969 $\pm$ 0.0064 & 0.5735 $\pm$ 0.0175 & 0.5121 $\pm$ 0.0357 & 0.5453 $\pm$ 0.0095 & 0.4842 & 0.611 \\
FBIngram:100 & Freebase & in-family / in-family & 0.4289 $\pm$ 0.0036 & 0.3816 $\pm$ 0.0039 & 0.3785 $\pm$ 0.0145 & 0.3462 $\pm$ 0.0051 & 0.2503 $\pm$ 0.0128 & 0.4383 & 0.449 \\
FBIngram:25 & Freebase & in-family / in-family & 0.3358 $\pm$ 0.0006 & 0.3065 $\pm$ 0.0041 & 0.2894 $\pm$ 0.0071 & 0.2675 $\pm$ 0.0046 & 0.2191 $\pm$ 0.0047 & 0.3830 & 0.388 \\
FBIngram:50 & Freebase & in-family / in-family & 0.2912 $\pm$ 0.0011 & 0.2633 $\pm$ 0.0034 & 0.2587 $\pm$ 0.0050 & 0.2430 $\pm$ 0.0049 & 0.2171 $\pm$ 0.0040 & 0.3303 & 0.338 \\
FBIngram:75 & Freebase & in-family / in-family & 0.3716 $\pm$ 0.0051 & 0.3352 $\pm$ 0.0059 & 0.3285 $\pm$ 0.0075 & 0.3082 $\pm$ 0.0034 & 0.2600 $\pm$ 0.0150 & 0.3905 & 0.403 \\
NLIngram:0 & NELL & new KG / new KG & 0.3589 $\pm$ 0.0106 & 0.3606 $\pm$ 0.0102 & 0.2886 $\pm$ 0.0045 & 0.2810 $\pm$ 0.0250 & 0.2359 $\pm$ 0.0104 & 0.3472 & 0.342 \\
NLIngram:100 & NELL & new KG / new KG & 0.4677 $\pm$ 0.0134 & 0.4346 $\pm$ 0.0014 & 0.4143 $\pm$ 0.0150 & 0.4148 $\pm$ 0.0063 & 0.2624 $\pm$ 0.0099 & 0.4422 & 0.471 \\
NLIngram:25 & NELL & new KG / new KG & 0.3534 $\pm$ 0.0082 & 0.3934 $\pm$ 0.0023 & 0.3114 $\pm$ 0.0037 & 0.3085 $\pm$ 0.0134 & 0.2680 $\pm$ 0.0495 & 0.3865 & 0.395 \\
NLIngram:50 & NELL & new KG / new KG & 0.4053 $\pm$ 0.0043 & 0.3677 $\pm$ 0.0148 & 0.3788 $\pm$ 0.0050 & 0.3552 $\pm$ 0.0067 & 0.2444 $\pm$ 0.0238 & 0.3976 & 0.407 \\
NLIngram:75 & NELL & new KG / new KG & 0.3504 $\pm$ 0.0024 & 0.3295 $\pm$ 0.0191 & 0.3330 $\pm$ 0.0028 & 0.3081 $\pm$ 0.0019 & 0.2244 $\pm$ 0.0096 & 0.3478 & 0.368 \\
WKIngram:100 & Wikidata & new KG / in-family & 0.1529 $\pm$ 0.0072 & 0.1494 $\pm$ 0.0107 & 0.1350 $\pm$ 0.0027 & 0.1105 $\pm$ 0.0008 & 0.0403 $\pm$ 0.0113 & 0.1782 & 0.164 \\
WKIngram:25 & Wikidata & new KG / in-family & 0.2825 $\pm$ 0.0084 & 0.2711 $\pm$ 0.0056 & 0.2868 $\pm$ 0.0117 & 0.2760 $\pm$ 0.0033 & 0.1468 $\pm$ 0.0169 & 0.3072 & 0.316 \\
WKIngram:50 & Wikidata & new KG / in-family & 0.1602 $\pm$ 0.0033 & 0.1578 $\pm$ 0.0069 & 0.1544 $\pm$ 0.0029 & 0.1560 $\pm$ 0.0039 & 0.1097 $\pm$ 0.0182 & 0.1577 & 0.166 \\
WKIngram:75 & Wikidata & new KG / in-family & 0.3521 $\pm$ 0.0152 & 0.3578 $\pm$ 0.0033 & 0.3164 $\pm$ 0.0026 & 0.3069 $\pm$ 0.0036 & 0.2137 $\pm$ 0.0192 & 0.3731 & 0.365 \\
FBNELL:FBNELL\_v1 & FBNELL & partial / partial & 0.1061 $\pm$ 0.0003 & 0.1024 $\pm$ 0.0010 & 0.0983 $\pm$ 0.0013 & 0.0959 $\pm$ 0.0002 & 0.0791 $\pm$ 0.0034 & 0.4734 & 0.485 \\
HM:1k & Hetionet-derived & new KG / new KG & 0.0568 $\pm$ 0.0045 & 0.0321 $\pm$ 0.0039 & 0.0464 $\pm$ 0.0026 & 0.0474 $\pm$ 0.0071 & 0.0176 $\pm$ 0.0019 & 0.0791 & 0.059 \\
HM:3k & Hetionet-derived & new KG / new KG & 0.0528 $\pm$ 0.0014 & 0.0266 $\pm$ 0.0047 & 0.0373 $\pm$ 0.0007 & 0.0315 $\pm$ 0.0004 & 0.0163 $\pm$ 0.0020 & 0.0634 & 0.037 \\
HM:5k & Hetionet-derived & new KG / new KG & 0.0468 $\pm$ 0.0017 & 0.0226 $\pm$ 0.0052 & 0.0343 $\pm$ 0.0008 & 0.0301 $\pm$ 0.0004 & 0.0158 $\pm$ 0.0015 & 0.0552 & 0.034 \\
ILPC2022:large & Wikidata & new KG / in-family & 0.2195 $\pm$ 0.0029 & 0.2018 $\pm$ 0.0212 & 0.1947 $\pm$ 0.0092 & 0.1470 $\pm$ 0.0100 & 0.0394 $\pm$ 0.0010 & 0.2971 & 0.290 \\
ILPC2022:small & Wikidata & new KG / in-family & 0.2283 $\pm$ 0.0055 & 0.2099 $\pm$ 0.0004 & 0.1969 $\pm$ 0.0012 & 0.1588 $\pm$ 0.0022 & 0.0683 $\pm$ 0.0036 & 0.2959 & 0.302 \\
Metafam:Metafam & Metafam & new KG / new KG & 0.3165 $\pm$ 0.0458 & 0.2161 $\pm$ 0.0292 & 0.4386 $\pm$ 0.0172 & 0.3819 $\pm$ 0.0732 & 0.2919 $\pm$ 0.0408 & 0.3298 & 0.238 \\
WikiTopicsMT1:health & Wikidata & new KG / in-family & 0.3140 $\pm$ 0.0043 & 0.1716 $\pm$ 0.0317 & 0.3089 $\pm$ 0.0008 & 0.2198 $\pm$ 0.0805 & 0.0954 $\pm$ 0.0051 & 0.2794 & 0.298 \\
WikiTopicsMT1:tax & Wikidata & new KG / in-family & 0.2099 $\pm$ 0.0186 & 0.0697 $\pm$ 0.0239 & 0.2107 $\pm$ 0.0122 & 0.2077 $\pm$ 0.0087 & 0.0476 $\pm$ 0.0127 & 0.2420 & 0.224 \\
WikiTopicsMT2:org & Wikidata & new KG / in-family & 0.0631 $\pm$ 0.0085 & 0.0535 $\pm$ 0.0025 & 0.0125 $\pm$ 0.0013 & 0.0146 $\pm$ 0.0003 & 0.0041 $\pm$ 0.0003 & 0.0832 & 0.095 \\
WikiTopicsMT2:sci & Wikidata & new KG / in-family & 0.2129 $\pm$ 0.0164 & 0.1122 $\pm$ 0.0218 & 0.2077 $\pm$ 0.0114 & 0.1951 $\pm$ 0.0097 & 0.0696 $\pm$ 0.0093 & 0.2576 & 0.258 \\
WikiTopicsMT3:art & Wikidata & new KG / in-family & 0.1721 $\pm$ 0.0102 & 0.1245 $\pm$ 0.0098 & 0.1254 $\pm$ 0.0008 & 0.1083 $\pm$ 0.0132 & 0.0247 $\pm$ 0.0073 & 0.2510 & 0.259 \\
WikiTopicsMT3:infra & Wikidata & new KG / in-family & 0.5275 $\pm$ 0.0105 & 0.5026 $\pm$ 0.0115 & 0.3675 $\pm$ 0.0177 & 0.3497 $\pm$ 0.0061 & 0.3289 $\pm$ 0.0022 & 0.6221 & 0.619 \\
WikiTopicsMT4:health & Wikidata & new KG / in-family & 0.6361 $\pm$ 0.0182 & 0.6348 $\pm$ 0.0072 & 0.5445 $\pm$ 0.0103 & 0.3098 $\pm$ 0.0201 & 0.4499 $\pm$ 0.0370 & 0.5565 & 0.624 \\
WikiTopicsMT4:sci & Wikidata & new KG / in-family & 0.2334 $\pm$ 0.0241 & 0.2338 $\pm$ 0.0151 & 0.1728 $\pm$ 0.0016 & 0.1355 $\pm$ 0.0071 & 0.0521 $\pm$ 0.0028 & 0.2934 & 0.274 \\
HM:indigo (excluded) & Hetionet-derived & new KG / new KG & — & — & — & — & — & 0.4361 & 0.440 \\
\bottomrule
\end{tabular}
}
\end{sidewaystable}

\begin{sidewaystable}
\centering
\scriptsize
\caption{the full 15 $\times$ 40 matrix (filtered MRR per seed), from \texttt{broad\_eval\_matrix.csv}.}
\label{tab:appC_per_seed}
\adjustbox{max width=0.92\textheight, max totalheight=0.9\textwidth}{%
\begin{tabular}{lrrrrr}
\toprule
split & GAT s0 / s1 / s2 & GINE (mean+max) s0 / s1 / s2 & GINE (sum) s0 / s1 / s2 & GraphSAGE s0 / s1 / s2 & R-GCN s0 / s1 / s2 \\
\midrule
FB15k237Inductive:v1 & 0.5238 / 0.5207 / 0.5327 & 0.4954 / 0.4943 / 0.5039 & 0.5171 / 0.5014 / 0.5036 & 0.4910 / 0.4986 / 0.5049 & 0.4569 / 0.4564 / 0.4609 \\
FB15k237Inductive:v2 & 0.5304 / 0.5314 / 0.5348 & 0.5110 / 0.5152 / 0.5215 & 0.5156 / 0.4947 / 0.4999 & 0.4886 / 0.4842 / 0.4874 & 0.4454 / 0.4397 / 0.4524 \\
FB15k237Inductive:v3 & 0.5063 / 0.5049 / 0.5043 & 0.4800 / 0.4754 / 0.4804 & 0.4789 / 0.4695 / 0.4705 & 0.4575 / 0.4566 / 0.4571 & 0.4221 / 0.4310 / 0.4141 \\
FB15k237Inductive:v4 & 0.4816 / 0.4848 / 0.4844 & 0.4600 / 0.4651 / 0.4619 & 0.4582 / 0.4432 / 0.4493 & 0.4353 / 0.4318 / 0.4386 & 0.3793 / 0.3854 / 0.3822 \\
NELLInductive:v1 & 0.6219 / 0.6152 / 0.6416 & 0.6180 / 0.6131 / 0.6345 & 0.5736 / 0.5999 / 0.5600 & 0.7610 / 0.7748 / 0.8061 & 0.3301 / 0.5546 / 0.5175 \\
NELLInductive:v2 & 0.5515 / 0.5650 / 0.5619 & 0.5163 / 0.5110 / 0.5253 & 0.4860 / 0.4764 / 0.4784 & 0.3452 / 0.3073 / 0.3184 & 0.3375 / 0.3679 / 0.3579 \\
NELLInductive:v3 & 0.5611 / 0.5581 / 0.5680 & 0.5430 / 0.5288 / 0.5381 & 0.5047 / 0.4947 / 0.4945 & 0.4159 / 0.3953 / 0.3927 & 0.3258 / 0.3570 / 0.3452 \\
NELLInductive:v4 & 0.5325 / 0.5353 / 0.5390 & 0.5123 / 0.4895 / 0.5067 & 0.4779 / 0.4613 / 0.4593 & 0.2793 / 0.2679 / 0.2470 & 0.2897 / 0.3292 / 0.3356 \\
WN18RRInductive:v1 & 0.6513 / 0.6639 / 0.6549 & 0.6605 / 0.6478 / 0.6592 & 0.6257 / 0.6108 / 0.6548 & 0.5209 / 0.5717 / 0.6041 & 0.6140 / 0.6013 / 0.5873 \\
WN18RRInductive:v2 & 0.6283 / 0.6391 / 0.6332 & 0.6406 / 0.6227 / 0.6386 & 0.6061 / 0.6038 / 0.6281 & 0.5624 / 0.5714 / 0.5958 & 0.6083 / 0.5992 / 0.5876 \\
WN18RRInductive:v3 & 0.3298 / 0.3300 / 0.3314 & 0.3139 / 0.3008 / 0.3171 & 0.3082 / 0.3105 / 0.3105 & 0.3073 / 0.3102 / 0.3110 & 0.2803 / 0.2813 / 0.2698 \\
WN18RRInductive:v4 & 0.5936 / 0.5986 / 0.6001 & 0.6012 / 0.5896 / 0.6000 & 0.5691 / 0.5586 / 0.5927 & 0.4768 / 0.5114 / 0.5482 & 0.5528 / 0.5485 / 0.5346 \\
FBIngram:100 & 0.4249 / 0.4318 / 0.4299 & 0.3852 / 0.3774 / 0.3823 & 0.3929 / 0.3640 / 0.3785 & 0.3405 / 0.3502 / 0.3478 & 0.2356 / 0.2593 / 0.2559 \\
FBIngram:25 & 0.3356 / 0.3364 / 0.3353 & 0.3021 / 0.3074 / 0.3101 & 0.2967 / 0.2826 / 0.2890 & 0.2728 / 0.2644 / 0.2653 & 0.2176 / 0.2244 / 0.2154 \\
FBIngram:50 & 0.2901 / 0.2911 / 0.2923 & 0.2617 / 0.2610 / 0.2672 & 0.2601 / 0.2531 / 0.2629 & 0.2459 / 0.2373 / 0.2457 & 0.2161 / 0.2215 / 0.2136 \\
FBIngram:75 & 0.3713 / 0.3768 / 0.3667 & 0.3372 / 0.3286 / 0.3398 & 0.3351 / 0.3203 / 0.3301 & 0.3044 / 0.3110 / 0.3093 & 0.2428 / 0.2673 / 0.2699 \\
NLIngram:0 & 0.3607 / 0.3476 / 0.3685 & 0.3684 / 0.3643 / 0.3491 & 0.2895 / 0.2926 / 0.2838 & 0.2635 / 0.2699 / 0.3096 & 0.2266 / 0.2341 / 0.2471 \\
NLIngram:100 & 0.4814 / 0.4547 / 0.4671 & 0.4360 / 0.4346 / 0.4333 & 0.4303 / 0.4006 / 0.4120 & 0.4188 / 0.4075 / 0.4181 & 0.2582 / 0.2552 / 0.2737 \\
NLIngram:25 & 0.3497 / 0.3476 / 0.3628 & 0.3959 / 0.3930 / 0.3913 & 0.3114 / 0.3151 / 0.3078 & 0.2956 / 0.3077 / 0.3223 & 0.2172 / 0.2708 / 0.3161 \\
NLIngram:50 & 0.4089 / 0.4066 / 0.4005 & 0.3821 / 0.3525 / 0.3686 & 0.3811 / 0.3731 / 0.3822 & 0.3585 / 0.3475 / 0.3595 & 0.2171 / 0.2555 / 0.2606 \\
NLIngram:75 & 0.3494 / 0.3487 / 0.3531 & 0.3409 / 0.3075 / 0.3402 & 0.3358 / 0.3328 / 0.3303 & 0.3096 / 0.3059 / 0.3088 & 0.2161 / 0.2223 / 0.2349 \\
WKIngram:100 & 0.1448 / 0.1586 / 0.1552 & 0.1588 / 0.1377 / 0.1517 & 0.1327 / 0.1342 / 0.1380 & 0.1109 / 0.1096 / 0.1109 & 0.0283 / 0.0418 / 0.0507 \\
WKIngram:25 & 0.2754 / 0.2918 / 0.2804 & 0.2748 / 0.2647 / 0.2739 & 0.2867 / 0.2752 / 0.2985 & 0.2724 / 0.2788 / 0.2769 & 0.1306 / 0.1455 / 0.1644 \\
WKIngram:50 & 0.1564 / 0.1628 / 0.1613 & 0.1641 / 0.1504 / 0.1588 & 0.1522 / 0.1533 / 0.1577 & 0.1521 / 0.1598 / 0.1561 & 0.0890 / 0.1169 / 0.1231 \\
WKIngram:75 & 0.3349 / 0.3635 / 0.3579 & 0.3595 / 0.3540 / 0.3600 & 0.3178 / 0.3134 / 0.3181 & 0.3110 / 0.3042 / 0.3056 & 0.1923 / 0.2194 / 0.2293 \\
FBNELL:FBNELL\_v1 & 0.1063 / 0.1057 / 0.1062 & 0.1024 / 0.1015 / 0.1034 & 0.0996 / 0.0981 / 0.0971 & 0.0958 / 0.0958 / 0.0961 & 0.0763 / 0.0829 / 0.0781 \\
HM:1k & 0.0554 / 0.0618 / 0.0532 & 0.0285 / 0.0315 / 0.0363 & 0.0469 / 0.0488 / 0.0436 & 0.0536 / 0.0491 / 0.0396 & 0.0154 / 0.0183 / 0.0190 \\
HM:3k & 0.0528 / 0.0541 / 0.0514 & 0.0213 / 0.0303 / 0.0282 & 0.0379 / 0.0366 / 0.0374 & 0.0319 / 0.0314 / 0.0311 & 0.0155 / 0.0148 / 0.0186 \\
HM:5k & 0.0485 / 0.0467 / 0.0451 & 0.0180 / 0.0283 / 0.0214 & 0.0352 / 0.0336 / 0.0340 & 0.0301 / 0.0298 / 0.0305 & 0.0151 / 0.0148 / 0.0175 \\
ILPC2022:large & 0.2163 / 0.2219 / 0.2202 & 0.1815 / 0.2001 / 0.2238 & 0.2012 / 0.1987 / 0.1841 & 0.1506 / 0.1357 / 0.1547 & 0.0396 / 0.0403 / 0.0384 \\
ILPC2022:small & 0.2221 / 0.2327 / 0.2301 & 0.2103 / 0.2096 / 0.2099 & 0.1982 / 0.1965 / 0.1959 & 0.1592 / 0.1564 / 0.1608 & 0.0646 / 0.0687 / 0.0717 \\
Metafam:Metafam & 0.3306 / 0.2653 / 0.3537 & 0.1828 / 0.2370 / 0.2286 & 0.4522 / 0.4192 / 0.4443 & 0.4234 / 0.4250 / 0.2974 & 0.3381 / 0.2608 / 0.2769 \\
WikiTopicsMT1:health & 0.3092 / 0.3173 / 0.3156 & 0.2078 / 0.1580 / 0.1489 & 0.3097 / 0.3082 / 0.3088 & 0.2678 / 0.1268 / 0.2647 & 0.0909 / 0.0944 / 0.1010 \\
WikiTopicsMT1:tax & 0.1915 / 0.2096 / 0.2286 & 0.0821 / 0.0422 / 0.0849 & 0.2244 / 0.2066 / 0.2011 & 0.2026 / 0.2028 / 0.2177 & 0.0619 / 0.0432 / 0.0377 \\
WikiTopicsMT2:org & 0.0534 / 0.0695 / 0.0664 & 0.0564 / 0.0519 / 0.0521 & 0.0110 / 0.0134 / 0.0131 & 0.0144 / 0.0150 / 0.0144 & 0.0039 / 0.0040 / 0.0045 \\
WikiTopicsMT2:sci & 0.1944 / 0.2187 / 0.2255 & 0.1252 / 0.0870 / 0.1244 & 0.2206 / 0.2037 / 0.1989 & 0.1928 / 0.1868 / 0.2057 & 0.0662 / 0.0801 / 0.0624 \\
WikiTopicsMT3:art & 0.1669 / 0.1839 / 0.1656 & 0.1194 / 0.1184 / 0.1358 & 0.1259 / 0.1245 / 0.1257 & 0.1234 / 0.1026 / 0.0988 & 0.0200 / 0.0210 / 0.0332 \\
WikiTopicsMT3:infra & 0.5209 / 0.5396 / 0.5221 & 0.4916 / 0.5146 / 0.5017 & 0.3878 / 0.3597 / 0.3550 & 0.3463 / 0.3461 / 0.3567 & 0.3289 / 0.3311 / 0.3268 \\
WikiTopicsMT4:health & 0.6151 / 0.6457 / 0.6475 & 0.6408 / 0.6268 / 0.6368 & 0.5517 / 0.5492 / 0.5327 & 0.3292 / 0.2890 / 0.3111 & 0.4072 / 0.4719 / 0.4705 \\
WikiTopicsMT4:sci & 0.2101 / 0.2582 / 0.2319 & 0.2472 / 0.2174 / 0.2369 & 0.1721 / 0.1716 / 0.1746 & 0.1369 / 0.1278 / 0.1418 & 0.0497 / 0.0552 / 0.0515 \\
\bottomrule
\end{tabular}
}
\end{sidewaystable}

\textbf{C.2 The 16-draw GAT portfolio (Section 5.1).} Sixteen trainings of
the frozen GAT recipe, differing only in seed, evaluated on the 25
GraIL and InGram benchmarks: 0.4318, 0.4346, 0.4367, 0.4221, 0.4331,
0.4248, 0.4302, 0.4354, 0.4310, 0.4293, 0.4310, 0.4293, 0.4217, 0.4247,
0.4367, 0.4286 (mean 0.4301, standard deviation 0.0048; ULTRA-3g
0.4299 on the same 25). The first three are the three seeds of the
paper. The portfolio is reported as a distribution: no draw was
selected or discarded.

\textbf{C.3 Alternative single training graphs} (Section 5.4): per-family $\Delta$ vs
the FB15k-237-Inductive-v1 reference. CoDEx-Small: 3 seeds,
gradient-step-matched to the reference (its 8$\times$ larger training set makes
second-matching meaningless; one of its gradient steps costs 13$\times$ the
reference's). ConceptNet100k and AristoV4 (100k--600k
facts) require sampled-subgraph training; their reference column is the
same model retrained in that regime, and these two columns are not
comparable to Table 2 or to the CoDEx-Small column. The preregistered
progress criteria (a gain above +0.02 MRR on a weak family without
losing more than 0.02 on the 40-graph mean) are met by no candidate.
From \texttt{results/monokg\_transfer.csv}.

\begin{table}[t]
\centering
\small
\caption{\textbf{alternative training graphs, per-family $\Delta$ MRR vs the reference of the same regime.}}

\begin{tabular}{lrrr}
\toprule
family & $\Delta$ CoDEx-Small & $\Delta$ ConceptNet100k & $\Delta$ AristoV4 \\
\midrule
Freebase & $-$0.140 & $-$0.194 & $-$0.096 \\
WordNet & $-$0.103 & $-$0.277 & $-$0.090 \\
NELL & $-$0.112 & $-$0.100 & $-$0.044 \\
Wikidata (target) & \textbf{$-$0.003} & $-$0.112 & $-$0.022 \\
Hetionet-derived (target) & $-$0.003 & +0.022 & $-$0.002 \\
Metafam & $-$0.023 & $-$0.240 & $-$0.239 \\
FBNELL & +0.251 & $-$0.137 & $-$0.075 \\
all-40 mean & $-$0.059 & $-$0.136 & $-$0.054 \\
\bottomrule
\end{tabular}
\end{table}

The FBNELL entry of the CoDEx-Small column (+0.251) is the
arbitrary-sign phenomenon of Section 5.3, not a repair.

\textbf{C.4 Rank dumps.} Every evaluation of this paper stored the filtered
rank of each query. The dumps behind Figure 3 (the fifteen models on
FBNELL) are released with the evaluation pipeline. For all fifteen
models no query lands at ranks 1--4 and the mode is rank 6; six queries
sit at rank 5 (seven for one GINE seed, a tie): the five whose answer is
one of the isolated entities and the one whose source is. ULTRA-3g's
ranks on the same queries (Figure 3b) show the contrast: it scores the
five isolated candidates identically on every query with a connected
source (maximal spread 1e-6) and places that block below a median of 184
of the 4,752 candidates; the five isolated-answer queries land at ranks
12, 30, 56, 156 and 530, and on 67 of the other 1,189 queries the block
still outranks the true answer.

\section{The KG$\rightarrow$RDB probe: protocol and per-key results}

\textbf{D.1 Databases and the graph actually probed.} The two databases are
RelBench's rel-f1 (Formula 1: 9 tables, 13 foreign-key columns) and
rel-stack (Stack Exchange: 7 tables, 11 foreign-key columns), about 57$\times$
larger (Table D1). Each is mapped by Step 1 and Step 2 of Section 3.1:
every row becomes an entity, every foreign-key column an anonymous
relation-type node, every non-null foreign-key cell a fact with its six
meta-relation edges. Cell values, column names, table names and
timestamps are discarded. The \texttt{CLASS} extension of Section 3.1 (one
entity-type node per table, linked to its rows by \texttt{is\_a}) was \emph{not}
exercised by the probe: the probed graph contains the three node kinds
of a knowledge graph and nothing else, and the pretrained models have
never seen a class node. Table membership enters the protocol in one
place only, the candidate set: a query on column \texttt{results.driverId} is
ranked against the rows of the \texttt{drivers} table. The probe therefore
tests the shared vocabulary in its knowledge-graph form (Step 1 + Step
2), and leaves the schema-level extension untested.

\begin{table}[htbp]
\centering
\scriptsize
\caption{the two databases (RelBench rel-f1 and rel-stack) as reified for the probe: rows $\rightarrow$ entities, FK columns $\rightarrow$ relation types, FK cells $\rightarrow$ facts (no class nodes; tables only constrain the candidate set). Reified nodes = rows + FK cells + FK columns; edges = 6 per FK cell.}
\label{tab:appD_db_sizes}
\adjustbox{max width=\textwidth}{%
\begin{tabular}{lrrrrrr}
\toprule
database & tables & rows & FK columns & FK cells & reified nodes & reified edges \\
\midrule
rel-f1 & 9 & 74,063 & 13 & 169,421 & 243,497 & 1,016,526 \\
rel-stack & 7 & 4,247,264 & 11 & 5,812,887 & 10,060,162 & 34,877,322 \\
\bottomrule
\end{tabular}
}
\end{table}

\textbf{D.2 Protocol.} For each foreign-key column in turn, its cells are
split at random (fixed seed) into a support half, which stays in the
inference graph, and a query half, which is removed from the graph and
asked as (row, column, ?) queries in the subject-to-object direction;
all other columns' cells remain in the graph. Candidates are the rows
of the referenced table. Ranking is filtered: every other known target
of the same (row, column) pair, from either half, is masked. Ties are
resolved pessimistically (a candidate scoring equal to the true answer
counts against it). In the paired runs of Tables D2 and D3, 60 queries
per column are drawn with a per-column seeded generator, identical for
every model and baseline. Each table reports Hits@10 and MRR with 95 \%
bootstrap confidence intervals over queries (1,000 resamples). Three
model-free references accompany each cell: the \emph{degree} heuristic,
which ranks candidates by how many support cells of the queried column
point to them, under the same filtering and tie rule; the \emph{random
baseline}, 1 / \#candidates; and, for the reified models, a
\emph{random-initialization control}, the same architecture with untrained
weights, which is at Hits@10 0.00 on every entity-referencing key of
both databases (Table D6) and is omitted from the paired tables. A key
is said to \emph{transfer} when the lower bound of the model's Hits@10
interval exceeds both the degree heuristic and the random-init control;
hits at Hits@10 $\le$ 0.10, all of which occur on event-referencing keys
(\texttt{*.raceId}, \texttt{postLinks.PostId}), are called \emph{marginal} and are not
counted in the text. Every number of this appendix was produced on the
single A100 of Section 4.

\textbf{D.3 Models.} \emph{The generic model} (``ours, 3 KGs'') is the memory-lean
GINE variant with mean+max aggregation used in Section 5 (there labeled
GINE (mean+max)), dimension 64, 12 layers, dropout 0.2, pretrained on the
transductive FB15k-237, WN18RR and CoDEx-Medium graphs (200, 100 and
100 batches per epoch respectively) with a recipe suited to that
corpus: constant learning rate $5\cdot 10^{-4}$, weight decay 0.01, edge dropout
0.15, gradient clipping at 1.0, batch size 8 with 4-step accumulation,
up to 40 epochs within a 6 h budget (22 completed), 423k parameters.
Unlike the models of Section 5 it has a \emph{query-conditioned readout}:
the candidate score is a function of both the candidate state and the
query relation state. Selection used the mean validation MRR of the
inductive v1 splits of FB15k-237, WN18RR and NELL-995 (validation
graphs distinct from the training graphs, never a test split). Three
pretraining seeds passed the sanity gate of Appendix B.6 (A = original
run, B and C = re-pretrainings); all three are re-ranked on the paired
draw in Tables D2--D5. \emph{``Ours, 1 KG, Section 5 recipe''} are the
three GINE (mean+max) checkpoints of Table 2, converted without
numerical change to the memory-lean implementation for the 10M-node
graph. \emph{``Ours, 1 KG, corpus recipe''} are three models trained with the
generic model's recipe on FB15k-237-Inductive v1 alone. \emph{``GAT §5''} are
the three GAT checkpoints of Table 2 (rel-f1 only; full-graph
propagation of the GAT does not fit the rel-stack graph in 80 GB).
\emph{ULTRA-3g/4g/50g} are the three released checkpoints, run zero-shot on
the raw foreign-key graph of Step 1 (rows as nodes, columns as relation
types, inverse relations added as in ULTRA's own data pipeline), with
the same queries, candidates, filtering and tie rule; an untrained ULTRA
of the same architecture served as its random-init control and stayed
at or below 0.03 Hits@10 on every key. Six raw hops were allowed on each side. ULTRA is
given the raw graph rather than the reified one because it degrades on
reified graphs \citep{huang2026hyperfoundationmodelinductive}; the honest
test of the baseline is the graph it was designed for.

\textbf{D.4 Paired results.} Tables D2 to D5 give the paired comparison on
the two databases: rel-f1 (Hits@10, then MRR), then rel-stack.

\begin{sidewaystable}
\centering
\scriptsize
\caption{rel-f1, paired probe — same 60 seeded queries per key for every column; Hits@10 [95\% bootstrap CI]. Degree = in-degree heuristic under the queried key. Three-seed columns list the seeds in order 0 / 1 / 2. The 3-KG column lists the three pretraining seeds of the generic model (A = original run, B and C = re-pretrainings with the same recipe), each re-ranked on the paired draw; CIs are omitted in this column, as in the other three-seed columns; Table D6 keeps their historical (unseeded) draws.}
\label{tab:appD_paired_f1_h10}
\adjustbox{max width=0.92\textheight, max totalheight=0.9\textwidth}{%
\begin{tabular}{lrrrrrrrr}
\toprule
key ($\rightarrow$ table, \#candidates) & degree & ULTRA-3g & ULTRA-4g & ULTRA-50g & GAT §5 (3 seeds) & ours, 1 KG, §5 recipe (3 seeds) & ours, 1 KG, corpus recipe (3 seeds) & ours, 3 KGs (seeds A / B / C) \\
\midrule
qualifying.raceId ($\rightarrow$ races, 820) & 0.00 & 0.00 [0.00, 0.00] & 0.10 [0.03, 0.18] & 0.07 [0.02, 0.13] & 0.08 / 0.05 / 0.08 & 0.05 / 0.05 / 0.03 & 0.07 / 0.05 / 0.07 & 0.08 / 0.08 / 0.12 \\
qualifying.driverId ($\rightarrow$ drivers, 857) & 0.35 & 0.53 [0.42, 0.67] & 0.75 [0.63, 0.85] & 0.65 [0.53, 0.77] & 0.00 / 0.00 / 0.00 & 0.00 / 0.00 / 0.00 & 0.00 / 0.00 / 0.00 & 0.93 / 0.93 / 0.97 \\
qualifying.constructorId ($\rightarrow$ constructors, 211) & 0.63 & 0.47 [0.35, 0.60] & 0.67 [0.55, 0.78] & 0.95 [0.90, 1.00] & 0.00 / 0.00 / 0.00 & 0.00 / 0.00 / 0.00 & 0.00 / 0.00 / 0.00 & 1.00 / 0.97 / 1.00 \\
races.circuitId ($\rightarrow$ circuits, 77) & 0.40 & 0.42 [0.30, 0.55] & 0.00 [0.00, 0.00] & 0.32 [0.20, 0.43] & 0.00 / 0.00 / 0.00 & 0.00 / 0.00 / 0.00 & 0.00 / 0.00 / 0.27 & 0.30 / 0.32 / 0.37 \\
standings.raceId ($\rightarrow$ races, 820) & 0.03 & 0.02 [0.00, 0.05] & 0.03 [0.00, 0.08] & 0.00 [0.00, 0.00] & 0.02 / 0.03 / 0.03 & 0.00 / 0.00 / 0.00 & 0.03 / 0.00 / 0.02 & 0.00 / 0.00 / 0.00 \\
standings.driverId ($\rightarrow$ drivers, 857) & 0.08 & 0.07 [0.02, 0.13] & 0.05 [0.00, 0.12] & 0.07 [0.02, 0.13] & 0.00 / 0.00 / 0.00 & 0.00 / 0.00 / 0.00 & 0.00 / 0.00 / 0.00 & 0.02 / 0.02 / 0.00 \\
constructor\_standings.raceId ($\rightarrow$ races, 820) & 0.00 & 0.02 [0.00, 0.05] & 0.02 [0.00, 0.05] & 0.02 [0.00, 0.05] & 0.00 / 0.00 / 0.00 & 0.00 / 0.00 / 0.00 & 0.00 / 0.02 / 0.00 & 0.02 / 0.02 / 0.02 \\
constructor\_standings.constructorId ($\rightarrow$ constructors, 211) & 0.42 & 0.42 [0.28, 0.53] & 0.40 [0.27, 0.52] & 0.38 [0.27, 0.52] & 0.00 / 0.00 / 0.00 & 0.00 / 0.00 / 0.00 & 0.00 / 0.00 / 0.00 & 0.08 / 0.47 / 0.00 \\
constructor\_results.raceId ($\rightarrow$ races, 820) & 0.00 & 0.00 [0.00, 0.00] & 0.00 [0.00, 0.00] & 0.02 [0.00, 0.05] & 0.00 / 0.00 / 0.00 & 0.00 / 0.00 / 0.00 & 0.03 / 0.00 / 0.02 & 0.02 / 0.02 / 0.02 \\
constructor\_results.constructorId ($\rightarrow$ constructors, 211) & 0.47 & 0.47 [0.35, 0.60] & 0.50 [0.38, 0.63] & 0.47 [0.35, 0.60] & 0.00 / 0.00 / 0.00 & 0.00 / 0.00 / 0.00 & 0.00 / 0.00 / 0.00 & 0.50 / 0.50 / 0.05 \\
results.raceId ($\rightarrow$ races, 820) & 0.00 & 0.03 [0.00, 0.08] & 0.03 [0.00, 0.08] & 0.08 [0.02, 0.15] & 0.00 / 0.00 / 0.00 & 0.10 / 0.00 / 0.07 & 0.02 / 0.00 / 0.03 & 0.08 / 0.08 / 0.08 \\
results.driverId ($\rightarrow$ drivers, 857) & 0.18 & 0.27 [0.17, 0.38] & 0.43 [0.30, 0.55] & 0.30 [0.18, 0.42] & 0.00 / 0.00 / 0.00 & 0.00 / 0.00 / 0.00 & 0.00 / 0.00 / 0.00 & 0.73 / 0.65 / 0.83 \\
results.constructorId ($\rightarrow$ constructors, 211) & 0.47 & 0.57 [0.45, 0.70] & 0.55 [0.42, 0.67] & 0.78 [0.68, 0.88] & 0.00 / 0.00 / 0.00 & 0.00 / 0.00 / 0.00 & 0.00 / 0.00 / 0.03 & 0.98 / 0.92 / 1.00 \\
\bottomrule
\end{tabular}
}
\end{sidewaystable}

\begin{sidewaystable}
\centering
\scriptsize
\caption{rel-f1, paired probe — same 60 seeded queries per key for every column; MRR. Degree = in-degree heuristic under the queried key. Three-seed columns list the seeds in order 0 / 1 / 2. The 3-KG column lists the three pretraining seeds of the generic model (A = original run, B and C = re-pretrainings with the same recipe), each re-ranked on the paired draw; Table D6 keeps their historical (unseeded) draws.}
\label{tab:appD_paired_f1_mrr}
\adjustbox{max width=0.92\textheight, max totalheight=0.9\textwidth}{%
\begin{tabular}{lrrrrrrrr}
\toprule
key ($\rightarrow$ table, \#candidates) & degree & ULTRA-3g & ULTRA-4g & ULTRA-50g & GAT §5 (3 seeds) & ours, 1 KG, §5 recipe (3 seeds) & ours, 1 KG, corpus recipe (3 seeds) & ours, 3 KGs (seeds A / B / C) \\
\midrule
qualifying.raceId ($\rightarrow$ races, 820) & 0.014 & 0.002 & 0.039 & 0.024 & 0.029 / 0.028 / 0.027 & 0.026 / 0.026 / 0.029 & 0.029 / 0.036 / 0.040 & 0.032 / 0.033 / 0.046 \\
qualifying.driverId ($\rightarrow$ drivers, 857) & 0.120 & 0.203 & 0.363 & 0.381 & 0.014 / 0.014 / 0.014 & 0.017 / 0.016 / 0.015 & 0.010 / 0.009 / 0.011 & 0.335 / 0.356 / 0.427 \\
qualifying.constructorId ($\rightarrow$ constructors, 211) & 0.162 & 0.223 & 0.324 & 0.715 & 0.047 / 0.047 / 0.048 & 0.068 / 0.053 / 0.052 & 0.037 / 0.041 / 0.039 & 0.544 / 0.472 / 0.832 \\
races.circuitId ($\rightarrow$ circuits, 77) & 0.193 & 0.194 & 0.019 & 0.141 & 0.033 / 0.033 / 0.033 & 0.031 / 0.028 / 0.027 & 0.031 / 0.031 / 0.094 & 0.101 / 0.147 / 0.109 \\
standings.raceId ($\rightarrow$ races, 820) & 0.028 & 0.007 & 0.011 & 0.010 & 0.014 / 0.013 / 0.014 & 0.003 / 0.005 / 0.009 & 0.015 / 0.005 / 0.005 & 0.022 / 0.022 / 0.013 \\
standings.driverId ($\rightarrow$ drivers, 857) & 0.049 & 0.040 & 0.029 & 0.029 & 0.008 / 0.008 / 0.008 & 0.004 / 0.004 / 0.002 & 0.002 / 0.002 / 0.003 & 0.029 / 0.036 / 0.005 \\
constructor\_standings.raceId ($\rightarrow$ races, 820) & 0.007 & 0.007 & 0.009 & 0.006 & 0.007 / 0.007 / 0.007 & 0.003 / 0.005 / 0.003 & 0.006 / 0.011 / 0.006 & 0.012 / 0.012 / 0.012 \\
constructor\_standings.constructorId ($\rightarrow$ constructors, 211) & 0.177 & 0.177 & 0.175 & 0.170 & 0.041 / 0.041 / 0.041 & 0.022 / 0.022 / 0.009 & 0.009 / 0.019 / 0.009 & 0.066 / 0.095 / 0.019 \\
constructor\_results.raceId ($\rightarrow$ races, 820) & 0.011 & 0.005 & 0.004 & 0.013 & 0.007 / 0.006 / 0.007 & 0.004 / 0.005 / 0.003 & 0.013 / 0.006 / 0.011 & 0.017 / 0.017 / 0.017 \\
constructor\_results.constructorId ($\rightarrow$ constructors, 211) & 0.244 & 0.246 & 0.198 & 0.217 & 0.043 / 0.041 / 0.043 & 0.019 / 0.015 / 0.013 & 0.012 / 0.015 / 0.013 & 0.100 / 0.102 / 0.058 \\
results.raceId ($\rightarrow$ races, 820) & 0.007 & 0.014 & 0.014 & 0.029 & 0.006 / 0.006 / 0.006 & 0.024 / 0.008 / 0.034 & 0.022 / 0.005 / 0.009 & 0.032 / 0.039 / 0.038 \\
results.driverId ($\rightarrow$ drivers, 857) & 0.079 & 0.124 & 0.186 & 0.151 & 0.009 / 0.009 / 0.009 & 0.006 / 0.004 / 0.009 & 0.005 / 0.003 / 0.007 & 0.231 / 0.190 / 0.257 \\
results.constructorId ($\rightarrow$ constructors, 211) & 0.197 & 0.243 & 0.280 & 0.453 & 0.030 / 0.030 / 0.030 & 0.022 / 0.012 / 0.015 & 0.015 / 0.009 / 0.037 & 0.534 / 0.427 / 0.703 \\
\bottomrule
\end{tabular}
}
\end{sidewaystable}

\begin{sidewaystable}
\centering
\scriptsize
\caption{rel-stack, paired probe — same 60 seeded queries per key for every column; Hits@10 [95\% bootstrap CI]. Degree = in-degree heuristic under the queried key. Three-seed columns list the seeds in order 0 / 1 / 2. The 3-KG column lists the three pretraining seeds of the generic model (A = original run, B and C = re-pretrainings with the same recipe), each re-ranked on the paired draw; CIs are omitted in this column, as in the other three-seed columns; Table D6 keeps their historical (unseeded) draws.}
\label{tab:appD_paired_stack_h10}
\adjustbox{max width=0.92\textheight, max totalheight=0.9\textwidth}{%
\begin{tabular}{lrrrrrrr}
\toprule
key ($\rightarrow$ table, \#candidates) & degree & ULTRA-3g & ULTRA-4g & ULTRA-50g & ours, 1 KG, §5 recipe (3 seeds) & ours, 1 KG, corpus recipe (3 seeds) & ours, 3 KGs (seeds A / B / C) \\
\midrule
votes.PostId ($\rightarrow$ posts, 333,893) & 0.00 & 0.00 [0.00, 0.00] & 0.00 [0.00, 0.00] & 0.00 [0.00, 0.00] & 0.00 / 0.00 / 0.00 & 0.00 / 0.00 / 0.00 & 0.00 / 0.00 / 0.00 \\
votes.UserId ($\rightarrow$ users, 255,360) & 0.13 & 0.10 [0.03, 0.18] & 0.55 [0.43, 0.67] & 0.73 [0.62, 0.83] & 0.00 / 0.00 / 0.00 & 0.00 / 0.00 / 0.02 & 0.32 / 0.42 / 0.35 \\
badges.UserId ($\rightarrow$ users, 255,360) & 0.07 & 0.05 [0.00, 0.12] & 0.00 [0.00, 0.00] & 0.00 [0.00, 0.00] & 0.00 / 0.00 / 0.00 & 0.00 / 0.00 / 0.00 & 0.00 / 0.00 / 0.00 \\
posts.OwnerUserId ($\rightarrow$ users, 255,360) & 0.05 & 0.13 [0.05, 0.22] & 0.83 [0.73, 0.92] & 0.02 [0.00, 0.05] & 0.00 / 0.00 / 0.00 & 0.00 / 0.02 / 0.83 & 0.83 / 0.83 / 0.98 \\
posts.ParentId ($\rightarrow$ posts, 333,893) & 0.00 & 0.00 [0.00, 0.00] & 0.02 [0.00, 0.07] & 0.27 [0.17, 0.38] & 0.00 / 0.00 / 0.00 & 0.00 / 0.00 / 0.00 & 0.00 / 0.00 / 0.00 \\
comments.UserId ($\rightarrow$ users, 255,360) & 0.18 & 0.17 [0.08, 0.27] & 0.17 [0.08, 0.27] & 0.23 [0.13, 0.35] & 0.00 / 0.00 / 0.00 & 0.00 / 0.00 / 0.32 & 0.62 / 0.70 / 0.63 \\
comments.PostId ($\rightarrow$ posts, 333,893) & 0.00 & 0.00 [0.00, 0.00] & 0.22 [0.12, 0.32] & 0.35 [0.23, 0.47] & 0.12 / 0.17 / 0.17 & 0.00 / 0.08 / 0.12 & 0.27 / 0.25 / 0.25 \\
postHistory.PostId ($\rightarrow$ posts, 333,893) & 0.00 & 0.00 [0.00, 0.00] & 0.32 [0.20, 0.43] & 0.32 [0.20, 0.43] & 0.33 / 0.35 / 0.33 & 0.00 / 0.10 / 0.08 & 0.32 / 0.30 / 0.28 \\
postHistory.UserId ($\rightarrow$ users, 255,360) & 0.20 & 0.17 [0.08, 0.27] & 0.15 [0.07, 0.23] & 0.22 [0.12, 0.33] & 0.00 / 0.00 / 0.00 & 0.00 / 0.00 / 0.75 & 0.82 / 0.85 / 0.83 \\
postLinks.PostId ($\rightarrow$ posts, 333,893) & 0.00 & 0.00 [0.00, 0.00] & 0.03 [0.00, 0.08] & 0.10 [0.03, 0.18] & 0.07 / 0.00 / 0.00 & 0.00 / 0.00 / 0.00 & 0.08 / 0.08 / 0.08 \\
postLinks.RelatedPostId ($\rightarrow$ posts, 333,893) & 0.03 & 0.02 [0.00, 0.05] & 0.03 [0.00, 0.08] & 0.05 [0.00, 0.12] & 0.00 / 0.00 / 0.02 & 0.02 / 0.00 / 0.02 & 0.07 / 0.10 / 0.10 \\
\bottomrule
\end{tabular}
}
\end{sidewaystable}

\begin{sidewaystable}
\centering
\scriptsize
\caption{rel-stack, paired probe — same 60 seeded queries per key for every column; MRR. Degree = in-degree heuristic under the queried key. Three-seed columns list the seeds in order 0 / 1 / 2. The 3-KG column lists the three pretraining seeds of the generic model (A = original run, B and C = re-pretrainings with the same recipe), each re-ranked on the paired draw; Table D6 keeps their historical (unseeded) draws.}
\label{tab:appD_paired_stack_mrr}
\adjustbox{max width=0.92\textheight, max totalheight=0.9\textwidth}{%
\begin{tabular}{lrrrrrrr}
\toprule
key ($\rightarrow$ table, \#candidates) & degree & ULTRA-3g & ULTRA-4g & ULTRA-50g & ours, 1 KG, §5 recipe (3 seeds) & ours, 1 KG, corpus recipe (3 seeds) & ours, 3 KGs (seeds A / B / C) \\
\midrule
votes.PostId ($\rightarrow$ posts, 333,893) & 0.001 & 0.001 & 0.001 & 0.001 & 0.000 / 0.000 / 0.000 & 0.000 / 0.000 / 0.000 & 0.000 / 0.001 / 0.000 \\
votes.UserId ($\rightarrow$ users, 255,360) & 0.066 & 0.065 & 0.466 & 0.735 & 0.000 / 0.000 / 0.000 & 0.000 / 0.000 / 0.017 & 0.305 / 0.358 / 0.310 \\
badges.UserId ($\rightarrow$ users, 255,360) & 0.034 & 0.015 & 0.000 & 0.002 & 0.000 / 0.000 / 0.000 & 0.000 / 0.000 / 0.000 & 0.000 / 0.000 / 0.000 \\
posts.OwnerUserId ($\rightarrow$ users, 255,360) & 0.033 & 0.072 & 0.685 & 0.021 & 0.000 / 0.000 / 0.000 & 0.000 / 0.017 / 0.608 & 0.516 / 0.490 / 0.623 \\
posts.ParentId ($\rightarrow$ posts, 333,893) & 0.000 & 0.000 & 0.008 & 0.174 & 0.000 / 0.000 / 0.000 & 0.000 / 0.000 / 0.000 & 0.000 / 0.000 / 0.000 \\
comments.UserId ($\rightarrow$ users, 255,360) & 0.116 & 0.108 & 0.064 & 0.129 & 0.000 / 0.000 / 0.000 & 0.000 / 0.000 / 0.319 & 0.500 / 0.512 / 0.489 \\
comments.PostId ($\rightarrow$ posts, 333,893) & 0.000 & 0.001 & 0.129 & 0.213 & 0.110 / 0.145 / 0.135 & 0.000 / 0.062 / 0.090 & 0.191 / 0.182 / 0.186 \\
postHistory.PostId ($\rightarrow$ posts, 333,893) & 0.000 & 0.002 & 0.209 & 0.210 & 0.230 / 0.217 / 0.194 & 0.000 / 0.087 / 0.070 & 0.203 / 0.200 / 0.191 \\
postHistory.UserId ($\rightarrow$ users, 255,360) & 0.080 & 0.072 & 0.063 & 0.105 & 0.000 / 0.000 / 0.000 & 0.000 / 0.000 / 0.717 & 0.734 / 0.736 / 0.756 \\
postLinks.PostId ($\rightarrow$ posts, 333,893) & 0.000 & 0.000 & 0.007 & 0.048 & 0.027 / 0.000 / 0.002 & 0.000 / 0.000 / 0.000 & 0.060 / 0.060 / 0.060 \\
postLinks.RelatedPostId ($\rightarrow$ posts, 333,893) & 0.017 & 0.008 & 0.016 & 0.029 & 0.001 / 0.001 / 0.005 & 0.005 / 0.001 / 0.005 & 0.055 / 0.063 / 0.062 \\
\bottomrule
\end{tabular}
}
\end{sidewaystable}

Reading the tables with the rule of D.2: the generic 3-KG model
transfers on the four driver and constructor keys of the \texttt{qualifying}
and \texttt{results} tables of rel-f1 and on six rel-stack keys (the four
author keys \texttt{votes.UserId}, \texttt{posts.OwnerUserId}, \texttt{comments.UserId},
\texttt{postHistory.UserId}, and the two post keys \texttt{comments.PostId},
\texttt{postHistory.PostId}); the six one-KG models transfer on no
entity-referencing rel-f1 key; the Section 5 recipe finds the two post
keys of rel-stack for all three seeds and never an author key, while
the corpus recipe on one graph finds the three author keys (and
\texttt{comments.PostId}) for one seed in three, and nothing above the
marginal threshold for the other two. ULTRA-3g exceeds both controls on
one key (\texttt{qualifying.driverId}) of rel-f1 and none of rel-stack;
ULTRA-4g on 3 and 4 keys; ULTRA-50g on 5 and 5, of which 3 and 4 are
above the marginal threshold.

\textbf{D.5 Robustness to the pretraining seed.} Table D6 reports the three
passing pretraining seeds of the generic model on the historical query
draws of the original study. Those draws were made with an unseeded
subsampler, so they cannot be reconstructed exactly; the split (support
half versus query half, seed 0) and the candidate sets are identical by
construction, as the random-baseline column, which depends only on the
candidate set, confirms on all 24 keys. Table D6 keeps those historical
draws for the record; Tables D2--D5 re-rank all three seeds on the same
seeded draw (60 queries per key, both databases). On that paired draw
all three seeds transfer on the same strong keys (four on rel-f1, six
on rel-stack), with per-key differences of at most 0.10 Hits@10 on
rel-f1 and 0.15 on rel-stack. Two boundary keys of rel-f1
(\texttt{constructor\_standings.constructorId},
\texttt{constructor\_results.constructorId}), on which the generic model does
not beat the degree heuristic, vary with the seed (0.47/0.50 for seed B
against 0.00/0.05 for seed C, seed A 0.08/0.50) and are reported as
such.

\begin{sidewaystable}
\centering
\scriptsize
\caption{robustness to the pretraining seed of the generic 3-KG model, Hits@10 [95\% CI] on the historical (unseeded) query draws; n = queries per key for that column. Seed A is the original pretraining run, B and C two re-pretrainings with the same recipe; a fourth draw failed the pretraining sanity gate (Appendix B). Random-init = same architecture, untrained; degree = in-degree heuristic (both on draw A).}
\label{tab:appD_seed_robustness}
\adjustbox{max width=0.92\textheight, max totalheight=0.9\textwidth}{%
\begin{tabular}{lrrrrr}
\toprule
database · key & random-init (A) & degree (A) & seed A (original), n & seed B, n & seed C, n \\
\midrule
rel-f1 · qualifying.raceId & 0.05 & 0.02 & 0.03 [0.00, 0.08], 60 & 0.03 [0.01, 0.06], 200 & 0.10 [0.07, 0.15], 200 \\
rel-f1 · qualifying.driverId & 0.00 & 0.33 & 0.93 [0.87, 0.98], 60 & 0.91 [0.87, 0.95], 200 & 0.95 [0.93, 0.98], 200 \\
rel-f1 · qualifying.constructorId & 0.00 & 0.70 & 1.00 [1.00, 1.00], 60 & 0.99 [0.98, 1.00], 200 & 1.00 [1.00, 1.00], 200 \\
rel-f1 · races.circuitId & 0.00 & 0.63 & 0.33 [0.22, 0.45], 60 & 0.34 [0.28, 0.40], 200 & 0.40 [0.33, 0.47], 200 \\
rel-f1 · standings.raceId & 0.05 & 0.05 & 0.12 [0.05, 0.20], 60 & 0.07 [0.04, 0.10], 200 & 0.04 [0.02, 0.07], 200 \\
rel-f1 · standings.driverId & 0.00 & 0.08 & 0.02 [0.00, 0.05], 60 & 0.01 [0.00, 0.03], 200 & 0.00 [0.00, 0.00], 200 \\
rel-f1 · constructor\_standings.raceId & 0.00 & 0.00 & 0.00 [0.00, 0.00], 60 & 0.01 [0.00, 0.03], 200 & 0.01 [0.00, 0.03], 200 \\
rel-f1 · constructor\_standings.constructorId & 0.00 & 0.38 & 0.10 [0.03, 0.18], 60 & 0.51 [0.43, 0.57], 200 & 0.01 [0.00, 0.01], 200 \\
rel-f1 · constructor\_results.raceId & 0.02 & 0.00 & 0.03 [0.00, 0.08], 60 & 0.04 [0.01, 0.06], 200 & 0.04 [0.01, 0.06], 200 \\
rel-f1 · constructor\_results.constructorId & 0.00 & 0.57 & 0.60 [0.47, 0.72], 60 & 0.56 [0.49, 0.62], 200 & 0.04 [0.01, 0.06], 200 \\
rel-f1 · results.raceId & 0.05 & 0.02 & 0.05 [0.00, 0.12], 60 & 0.05 [0.02, 0.09], 200 & 0.08 [0.04, 0.12], 200 \\
rel-f1 · results.driverId & 0.00 & 0.08 & 0.78 [0.68, 0.88], 60 & 0.67 [0.60, 0.73], 200 & 0.81 [0.76, 0.86], 200 \\
rel-f1 · results.constructorId & 0.00 & 0.43 & 0.97 [0.92, 1.00], 60 & 0.91 [0.87, 0.95], 200 & 0.98 [0.96, 1.00], 200 \\
rel-stack · votes.PostId & 0.00 & 0.00 & 0.00 [0.00, 0.00], 60 & 0.00 [0.00, 0.00], 60 & 0.02 [0.00, 0.07], 60 \\
rel-stack · votes.UserId & 0.00 & 0.12 & 0.47 [0.35, 0.58], 60 & 0.60 [0.48, 0.72], 60 & 0.47 [0.35, 0.58], 60 \\
rel-stack · badges.UserId & 0.00 & 0.03 & 0.00 [0.00, 0.00], 60 & 0.00 [0.00, 0.00], 60 & 0.00 [0.00, 0.00], 60 \\
rel-stack · posts.OwnerUserId & 0.00 & 0.03 & 0.75 [0.65, 0.85], 60 & 0.75 [0.65, 0.85], 60 & 0.93 [0.87, 0.98], 60 \\
rel-stack · posts.ParentId & 0.00 & 0.00 & 0.00 [0.00, 0.00], 60 & 0.00 [0.00, 0.00], 60 & 0.00 [0.00, 0.00], 60 \\
rel-stack · comments.UserId & 0.00 & 0.18 & 0.58 [0.47, 0.70], 60 & 0.70 [0.58, 0.82], 60 & 0.55 [0.43, 0.68], 60 \\
rel-stack · comments.PostId & 0.00 & 0.00 & 0.25 [0.15, 0.35], 60 & 0.25 [0.15, 0.35], 60 & 0.25 [0.15, 0.35], 60 \\
rel-stack · postHistory.PostId & 0.00 & 0.00 & 0.48 [0.35, 0.62], 60 & 0.48 [0.35, 0.62], 60 & 0.43 [0.30, 0.57], 60 \\
rel-stack · postHistory.UserId & 0.00 & 0.22 & 0.77 [0.67, 0.87], 60 & 0.80 [0.70, 0.90], 60 & 0.83 [0.73, 0.92], 60 \\
rel-stack · postLinks.PostId & 0.00 & 0.00 & 0.08 [0.02, 0.17], 60 & 0.08 [0.02, 0.17], 60 & 0.05 [0.00, 0.12], 60 \\
rel-stack · postLinks.RelatedPostId & 0.03 & 0.03 & 0.03 [0.00, 0.08], 60 & 0.05 [0.00, 0.12], 60 & 0.03 [0.00, 0.08], 60 \\
\bottomrule
\end{tabular}
}
\end{sidewaystable}

\textbf{D.6 Failure profile.} The failures of the reified models are the
same on both databases and for every seed: keys whose targets form a
large class of near-interchangeable, time-indexed events (the five
\texttt{*.raceId} keys of rel-f1; \texttt{votes.PostId}, \texttt{posts.ParentId} and the two
\texttt{postLinks.*} keys of rel-stack) stay near random, and so does
\texttt{badges.UserId}, the one key whose subject rows carry no other foreign
key and therefore no compositional context. The keys that transfer are
those where the queried fact sits in a dense co-key context (a
\texttt{results} row also names its race and its constructor) and the target
class is a set of persistent, structurally rich entities (drivers,
constructors, users). ULTRA-50g does not share this profile entirely:
it reads \texttt{posts.ParentId} (0.27) where the reified models are at 0.00,
and it is at 0.02 on \texttt{posts.OwnerUserId} where they are at 0.83.

\clearpage
\bibliographystyle{plainnat}
\bibliography{references}

\end{document}